\documentclass[journal]{IEEEtran}
\usepackage{color}
\usepackage{adjustbox}
\usepackage{multirow}
\usepackage{booktabs}
\usepackage{subfigure}
\usepackage{graphicx,multicol,subfig}
\usepackage{threeparttable}
\usepackage{mathtools}
\usepackage{amsmath}
\usepackage{nicefrac}

\def\eg{\textit{e.g}. } 
\def\ie{\textit{i.e}. } 
\def\cf{\textit{c.f}. }

\def\wrt{w.r.t. } \def\etal{\textit{et al}. }

\begin{document}

%\title{An Attack on Feature Level-based\\ Facial Soft-biometric Privacy Enhancement}
\title{An Attack on Facial Soft-biometric\\ Privacy Enhancement}
%x
%
% author names and IEEE memberships
% note positions of commas and nonbreaking spaces ( ~ ) LaTeX will not break
% a structure at a ~ so this keeps an author's name from being broken across
% two lines.
% use \thanks{} to gain access to the first footnote area
% a separate \thanks must be used for each paragraph as LaTeX2e's \thanks
% was not built to handle multiple paragraphs
%

 \author{D. Osorio-Roig,
         C. Rathgeb,
         P. Drozdowski, P. Terhörst, V. Štruc,
         and~C. Busch%,~\IEEEmembership{Member,~IEEE}% <-this % stops a space
 \thanks{D. Osorio-Roig, C. Rathgeb, P. Drozdowski and~C. Busch are with the Department of Computer Science, Hochschule Darmstadt, Darmstadt, Germany. \\
  P. Terhörst is with the Fraunhofer
Institute for Computer Graphics Research, Darmstadt, Germany, and Department of Computer Science, Norwegian University of Science and Technology, Gj{\o}vik, Norway. \\
 V. Štruc is with Faculty of Electrical Engineering,
University of Ljubljana, Ljubljana, Slovenia.\protect\\
E-mail: daile.osorio-roig@h-da.de}% <-this % stops a space
 }

\IEEEtitleabstractindextext{
\begin{abstract}
In the recent past, different researchers have proposed  privacy-enhancing face recognition systems designed to conceal soft-biometric attributes at feature level. These works have reported impressive results, but generally did not consider specific attacks in their analysis of privacy protection. %In most cases, the privacy protection capabilities of these schemes are tested through simple machine learning-based classifiers and visualisations of dimensionality reduction tools.
We introduce an attack on said schemes based on two observations: (1) highly similar facial representations usually originate from face images with similar soft-biometric attributes; (2) to achieve high recognition accuracy, robustness against intra-class variations within facial representations has to be retained in their privacy-enhanced versions. The presented attack only requires the privacy-enhancing algorithm as a black-box and a relatively small database of  face images with annotated soft-biometric attributes. Firstly, an intercepted privacy-enhanced face representation is compared against the attacker's database. Subsequently, the unknown attribute is inferred from the attributes associated with the highest obtained similarity scores. In the experiments, the attack is applied against two state-of-the-art approaches. The attack is shown to circumvent the privacy enhancement to a considerable degree and is able to correctly classify gender with an accuracy of up to approximately 90\%. Future works on privacy-enhancing face recognition are encouraged to include the proposed attack in evaluations on the privacy protection.
\end{abstract}

% Note that keywords are not normally used for peerreview papers.
}
% For peer review papers, you can put extra information on the cover
% page as needed:
% \ifCLASSOPTIONpeerreview
% \begin{center} \bfseries EDICS Category: 3-BBND \end{center}
% \fi
%
% For peerreview papers, this IEEEtran command inserts a page break and
% creates the second title. It will be ignored for other modes.
\maketitle

\IEEEdisplaynontitleabstractindextext
\IEEEpeerreviewmaketitle

\section{Introduction}\label{sec:introduction}
\IEEEPARstart{F}{ace} recognition technologies are deployed in many personal, commercial, and governmental identity management systems around the world. Current state-of-the-art face recognition technologies utilise deep learning and massive training datasets to embed face images as discriminative representations in the latent space \cite{Ranjan18a,Guo-DeepFaceSurvey-2019}. Similar kinds of deep learning techniques, \eg deconvolutional neural networks, have shown impressive results for reconstructing facial images from their corresponding embeddings \cite{Mai-Reconstruction-2019}. Further, it has been demonstrated that, sensitive soft-biometric information, \eg gender, race, or age, can be directly derived from facial embeddings \cite{9304874,Terhoerst-Soft-Biometric-2021}.
 
In response to these privacy issues, a considerable amount of research has been conducted over the past years. In order to protect individuals' privacy, \emph{biometric template protection} schemes have been proposed for various biometric characteristics, including the face. Biometric template protection methods are commonly categorised as \emph{cancelable biometrics} and \emph{biometric cryptosystems}. Cancelable biometrics employ transforms in the signal or feature domain which enable a biometric comparison in the transformed domain \cite{Patel-CancelableBiometrics-2015}. In contrast, the majority of biometric cryptosystems binds a key to a biometric feature vector resulting in a protected template. Biometric authentication is then performed indirectly by verifying the correctness of a retrieved key \cite{BUludag04a}. For comprehensive surveys on this topic, the interested reader is referred to \cite{BJain08b,Rathgeb11e}. Alternatively, homomorphic encryption has frequently been suggested for biometric template protection \cite{Aguilar-Homomorphic-2013}. Homomorphic encryption makes it possible to compute operations in the encrypted domain which are functionally equivalent to those in the plaintext domain and thus enables the estimation of certain distances between protected biometric templates. Biometric template protection are designed to fulfill the major requirements of irreversibility and unlinkability which are defined in ISO/IEC IS 24745 \cite{ISO11-TemplateProtection}.

\begin{figure}[!t]
    \centering
    \subfigure[Original (unprotected) ]{\includegraphics[width=\linewidth]{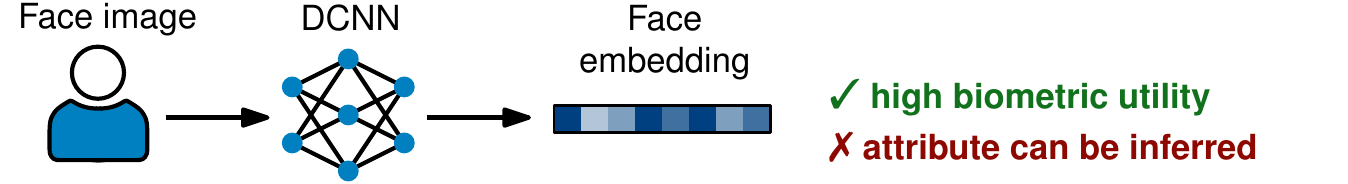}}\vspace{0.2cm}
    \subfigure[Privacy-enhanced (protected)]{\includegraphics[width=\linewidth]{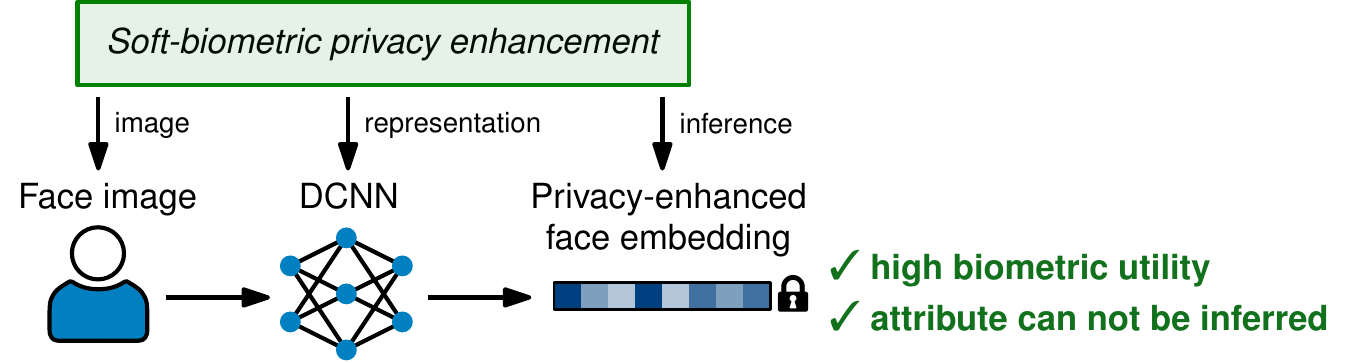}}
    \caption{Extraction of facial embeddings: (a) original face embeddings exhibit high biometric utility, but sensitive attributes can be derived from them; (b) application of soft-biometric privacy enhancement at image, representation or inference level is usually claimed to result in face embeddings with high biometric utility of which sensitive attributes can not be inferred.}
    \label{fig:privacy_enhancement_overview}\vspace{-0.3cm}
\end{figure}

In addition to face-based biometric template protection, methods that attempt to remove (or conceal) certain sensitive information from facial biometric data (while leaving other useful information unchanged) have been proposed by various research laboratories. Said schemes have recently been summarised under the umbrella term \emph{privacy–enhancing face biometrics}, a comprehensive survey can be found in \cite{Meden21a}. A large amount of published methods which are referred to as \emph{soft-biometric privacy enhancement} aim at removing or suppressing sensitive attributes in facial data. In the context of a face recognition system, this group of techniques can be applied at either image level, representation level, or at inference level \cite{Meden21a}. Approaches applied on image level, \eg obfuscation, have been shown to enhance privacy at the cost of biometric utility. In other words, privacy-enhanced face images obtained by said techniques become less usable for facial recognition tasks. Further, different methods have been applied at representation level or inference level, \ie these methods operate at feature level. Interestingly, the latter schemes have been reported to retain biometric utility and, at the same time, provide strong privacy protection \cite{Meden21a}, see Figure~\ref{fig:privacy_enhancement_overview}. This clearly contradicts with the assumption that a removal or suppression of facial information yield less discriminative face embeddings which results in a decrease in biometric performance, analogous to methods applied on image level. This necessitates a closer examination of soft-biometric privacy enhancement methods. In particular since published approaches often lack a rigorous analysis with respect to privacy protection  \cite{TerhoerstBIOSIG}. 

\begin{table*}[!t]
	\scriptsize
	\centering
	\caption{Overview of relevant facial soft-biometric privacy enhancement approaches operating at feature level (results reported for best configurations; note the differences in the used evaluation datasets and performance metrics).}
	\label{tab:related-works}
    	\begin{adjustbox}{max width=\linewidth}
	\begin{threeparttable}

	\begin{tabular}{ccccc cc cc}
	\toprule
	   \multirow{2}{*}{\textbf{Authors}}& \multirow{2}{*}{\textbf{Method}}&\multirow{2}{*}{\textbf{Level}}&\multirow{2}{*}{\textbf{Protected Attribute}}&\multirow{2}{*}{\textbf{Datasets} }           &\multicolumn{2}{c}{\textbf{Original}}&\multicolumn{2}{c}{\textbf{Privacy-Enhanced}} \\ \cmidrule{6-9}
	                
	                   & &                  &                            &               &       \textbf{Biometric Perf.}&\textbf{Classification Perf.} &  \textbf{Biometric Perf.}&\textbf{Classification Perf.} \\ \midrule
	                    
	  \multirow{2}{*}{Terh{\"o}rst \etal \cite{Terhorst-NoiseTransformation-2019}}& \multirow{2}{*}{CSN-Transformation}  &\multirow{2}{*}{Representation}&Gender&\multirow{2}{*}{ColorFeret}&\multirow{2}{*}{$\sim{0.09\%}$ EER}&$\sim{90.0\%}$ ADA&\multirow{2}{*}{$\sim{0.18\%}$ EER} &$\sim{65.00\%}$ ADA \\ 
	  
	                                           &                                & &Age& &&--&& 10.6\% ASR \\ \midrule
	  
	  \multirow{2}{*}{Morales \etal \cite{Morales-SensitiveNets-2020}} & \multirow{2}{*}{SensitiveNets} & \multirow{2}{*}{Representation}&Gender&\multirow{2}{*}{LFW}&\multirow{2}{*}{98.4\% VA }&97.70\% ADA
	  &\multirow{2}{*}{95.8\% VA}&54.6\% ADA \\
	                      & &         &Ethnicity&                         &               &98.8\% ADA   & &53.5\% ADA\\ \midrule
	   
	   \multirow{2}{*}{Terh{\"o}rst \etal \cite{Terhorst-IncrementalElimimination-2019}}& \multirow{2}{*}{IVE} &\multirow{2}{*}{Representation}&Gender&\multirow{2}{*}{ColorFeret}&\multirow{2}{*}{3.1\% EER} & 94.8\% COCR&\multirow{2}{*}{3.8\% EER}& 77.9\% COCR\\ 
	   
	                    & &    &Age       &   &   &68.7\% COCR    &       &50.6\% COCR \\ \midrule
	   
	   \multirow{3}{*}{Bortolato \etal \cite{Bortolato-FeatureDisentanglement-2020}}& \multirow{3}{*}{PFRNet}& \multirow{3}{*}{Representation}& \multirow{3}{*}{Gender}&CelebA&5.9\% EER&1.8\% $fic$ &8.6\% EER &43.5\% $fic$ \\
	   
	                                                                            &    &                                   &
	                                                                                &Adience&5.6\% EER&14.5\% $fic$ & 6.4\% EER & 50.2\% $fic$   \\
	                                                                              &  &                                   &                           &LFW&1.8\% EER&4.9\% $fic$&2.8\% EER&41.4\% $fic$ \\ \midrule
	   \multirow{3}{*}{ Terh{\"o}rst \etal \cite{Terhorst-SoftbiometricEnhancingPE-MIU-2020}}& \multirow{3}{*}{PE-MIU} &\multirow{3}{*}{Inference}&\multirow{3}{*}{Gender}&LFW&0.49\% EER&89.50\% ADA&0.56\% EER&50.23\% ADA\\
	                                                                                
	                                                                             &       &           &      &Adience&3.27\% EER &89.81\% ADA &3.63\% EER&44.71\% ADA\\
	                                                                     &               &           &      &ColorFeret&2.15\% EER& 97.62\% ADA&3.11\% EER& 51.87 ADA\\ \midrule
	                                                                                    
	   \multirow{5}{*}{Terh{\"o}rst \etal \cite{Terhorst-NegativeFaceRecognition-2020}} & \multirow{5}{*}{NFR} &\multirow{5}{*}{Inference}&Gender&\multirow{3}{*}{ColorFeret}&\multirow{3}{*}{1.97\% EER}&97.30\% ADA &\multirow{3}{*}{3.18\%  EER}&22.2\% ASR \\
	   
	                                              &  &                           &Age&       &&57.40\% ADA& &30.2\% ASR \\
	                                             &   &                           &Ethnicity&      &&88.73\% ADA& &14.6\% ASR \\ 
	                                                
	                                           &     &                          &Gender&\multirow{2}{*}{Adience}&\multirow{2}{*}{3.83\% EER}&84.91\% ADA  &\multirow{2}{*}{4.43\% EER}&26.1\% ASR \\
	                                                
	                                            &    & &Age& & &60.36\% ADA &&28.7\% ASR \\

	\bottomrule
	\end{tabular}
	\begin{tablenotes}
	\item EER: Equal Error Rate, VA: Verification Accuracy, ADA: Attribute Decision Accuracy, ASR: Attribute Suppression Rate, COCR: Correct Overall Classification Rate, $fic$: Fraction of Incorrectly Classified Images

	\end{tablenotes}
	\end{threeparttable}
	\end{adjustbox}
\end{table*}

The main contribution of this work is the proposal of a novel attack on  privacy-enhancing face recognition systems. Here, we mainly focus on methods operating at representation or inference level while the attack is generally applicable to any soft-biometric privacy enhancement method (including image level-based methods). The attack builds upon the following observations: it has recently been shown that facial recognition algorithms produce higher similarity scores and, hence, significantly more false matches for subjects with similar soft-biometric attributes -- in particular gender and race. This effect is referred to as \emph{broad homogeneity} \cite{Howard-DemographicEffectsFace-2019}. Further, it has been shown that it is possible that face recognition algorithms operate on facial features that are unrelated to soft-biometric attributes, albeit with somewhat lower recognition accuracy \cite{howard2020quantifying}.

We show that the aforementioned properties also hold for privacy-enhancing face recognition systems. This can be exploited to attack these schemes, \ie infer soft-biometric attributes from privacy-enhanced face embeddings. In the proposed attack, a face database with known soft-biometric attributes is used to generate a set of privacy-enhanced face representations against which a privacy-enhanced face representation with unknown soft-biometric attributes is compared. The best obtained similarity scores are then analysed to derive the unknown attributes of the attacked privacy-enhanced face representation.   The attack can be performed offline and only requires the privacy-enhancing algorithm as black box and an arbitrary set of facial images with known soft-biometric attributes. In experimental evaluations, the attack is applied to privacy-enhanced face representation obtained by two recently published algorithms, \ie privacy-enhancing face-representation learning network (PFRNet) \cite{Bortolato-FeatureDisentanglement-2020} and privacy-enhancing face recognition based on minimum information units (PE-MIU) \cite{Terhorst-SoftbiometricEnhancingPE-MIU-2020}. High success rates of up to 90\% with respect to gender prediction are obtained for attacking both state-of-the-art algorithms. 

The results reported in this work indicate that privacy protection capabilities of  facial soft–biometric privacy enhancement methods are commonly over-estimated in the current scientific literature. Towards the creation of privacy-preserving biometric systems various attacks have been proposed against different types of popular biometric cryptosystems and cancelable biometrics, \eg in \cite{KONG20061359,4430534}. Uncovered gaps in privacy protection have in turn led to (continuous) improvements of such schemes. Therefore, we believe that the developments of facial soft–biometric privacy enhancement can benefit from considering the proposed attack. In particular, to advance developments of facial soft–biometric privacy enhancement, it is strongly suggested to employ the proposed kind of attack in evaluations of privacy protection capabilities of future methods.

This work is organised as follows: section~\ref{sec:works} briefly summarises most relevant works on soft-biometric privacy-enhancing techniques applied at feature level. Section~\ref{sec:attacks} describes the proposed attack in detail. The experimental setup and results are reported in sections~\ref{sec:experimental} and~\ref{sec:experimental-results}, respectively. They are subsequently discussed in section~\ref{sec:discussion}, while section~\ref{sec:conclusion} contains a summary and concluding remarks.

\section{Related works}
\label{sec:works}
Several efforts have been made in recent years to introduce different soft-biometric privacy-enhancing techniques at feature level, \ie approaches operating at representation or inference level. Table~\ref{tab:related-works} lists the most relevant works in this research area. The performance metrics are reported in the table exactly as in the cited papers. Note that differently named metrics often correspond to the same underlying concept, \eg ADA is expected to be the same as COCR. 

Terh{\"o}rst \etal \cite{Terhorst-NoiseTransformation-2019} proposed a Cosine–Sensitive Noise (CSN) transformation applied to face embeddings to enhance privacy in terms of gender and age attributes. To this end, the authors introduced an specific type of noise over the face representation which hides the soft–biometric information.
Morales \etal \cite{Morales-SensitiveNets-2020} proposed SensitiveNets, a privacy-preserving learning method. By incorporating soft-biometric classifiers in the loss function of during algorithm training, this approach learns new feature representations suppressing gender and ethnicity information. 
Terh{\"o}rst \etal \cite{Terhorst-IncrementalElimimination-2019} proposed a strategy called Incremental Variable
Elimination (IVE) to eliminate (or remove) components related to soft-biometric information from the face feature representation. Bortolato \etal \cite{Bortolato-FeatureDisentanglement-2020} managed to learn a disentangled feature representation in their so-called Privacy-Enhancing Face-Representation learning Network (PFRNet). PFRNet is an autoencoder which learns to separate gender attributes from the identity information.  

A few works operating at the inference level have been proposed recently. These methods apply transformations and adapt the biometric comparator accordingly.  Terh{\"o}rst \etal \cite{Terhorst-NegativeFaceRecognition-2020} proposed such a method based on Negative Face Recognition (NFR). So-called negative embeddings are obtained by introducing features to them that are intentionally different from the original (positive) embeddings, thereby concealing soft-biometric attributes. %Accordingly, a particular comparison process based on dissimilarities is also proposed.
Further, Terh{\"o}rst \etal \cite{Terhorst-SoftbiometricEnhancingPE-MIU-2020} proposed the Privacy-Enhancing face recognition approach based on Minimum Information Units (PE-MIU). This method allows the creation privacy-enhanced face template by partitioning the original feature vector into smaller parts (called minimum information units). Then, these blocks are randomly shuffled to obtain a privacy-enhanced template. 

Whereas several authors have explored the development of novel techniques for removal of information on soft-biometrics with promising results, there still exists a need for deeper analysis of the achieved privacy protection. Terh{\"o}rst~\etal~\cite{TerhoerstBIOSIG} recently argued that the absence of a standardized evaluation
protocol hampers a meaningful comparison of proposed approaches to facial soft-biometric privacy enhancement. They propose a framework to evaluate the trade-off between suppressing an attribute and maintaining
the recognition performance. However, their framework does not consider specific attacks. 

\begin{figure*}[!ht]
    \centering
    \includegraphics[width=\textwidth]{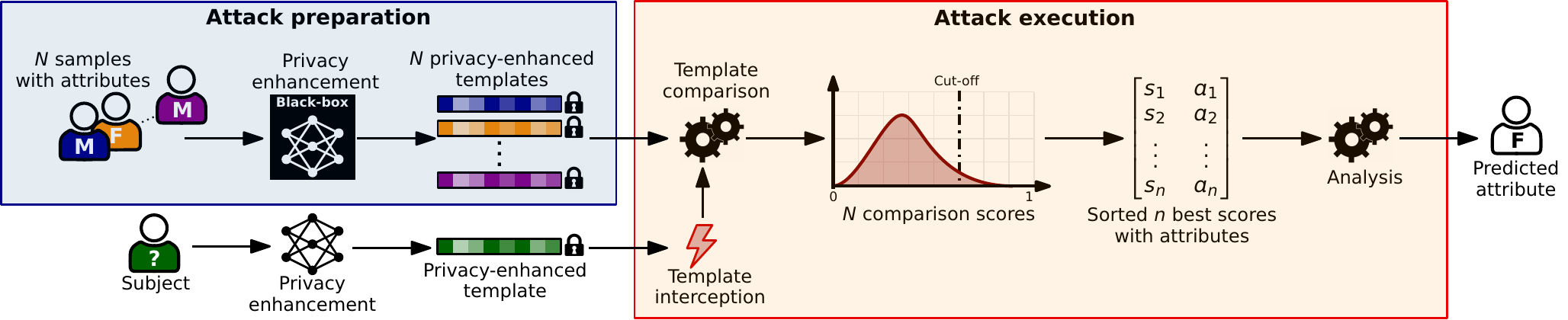}
    \caption{Overview of the attack: an attacker is in possession of the soft-biometric privacy enhancement method and applies it to a database of images with known labels (collected in the preparation phase); then, an intercepted privacy enhanced face embedding is compared against the database and the best scores are analysed to predict the soft-biometric attribute.}
    \label{fig:overview}
\end{figure*}
\section{Proposed Attack}
\label{sec:attacks}
This section presents the proposed attack. Subsection \ref{sec:attack-background} provides background information and theoretical foundations of the attack. Figure~\ref{fig:overview} shows an overview of the proposed attack; a detailed description of the attack execution is given in subsection \ref{sec:attack-analysis}.

\subsection{Background}
\label{sec:attack-background}
The proposed attack relies on several observations about:

\begin{enumerate}
    \item The effects of broad homogeneity and demographic differentials in face recognition.
    \item Properties and general operating principles of the privacy-enhancing methods the attack is aimed at.
\end{enumerate}

Regarding the first of the above, let $P$ denote a probability measure and $\mathrm{s}$ a similarity scoring function between two non-mated samples with given soft-biometric attributes $a_{1}$ and $a_{2}$, which can be identical (\eg female vs female) or different (\eg female vs male). For the purpose of this example, let $\mathrm{s}$ return similarity scores in the range $[0,1]$, where $0$ represents a complete dissimilarity and $1$ a perfect similarity. 

Many works have shown that comparisons between non-mated samples of same/similar soft-biometric attributes tend to generally yield higher similarity scores and consequently more frequent false matches, \eg in \cite{Grother-NIST-FRVTBias-2019,Howard-DemographicEffectsFace-2019,Sirotin-FairnessIdentification-2021}. This property is especially pertinent in face recognition, but does not necessarily hold for all other biometric characteristics (\eg iris) \cite{howard2020quantifying}. In many face recognition systems, the relation, 
\begin{equation}
    \label{eq:prob_score}
     P(\mathrm{s}_{{a_{1}} = a_{2}}(\cdot) > \mathrm{s}_{a_{1} \neq a_{2}}(\cdot)) \gg P(\mathrm{s}_{a_{1} = a_{2}}(\cdot) < \mathrm{s}_{a_{1} \neq a_{2}}(\cdot))
\end{equation}generally holds true, where $\mathrm{s}_{{a_{1}} = a_{2}}(\cdot)$ and $\mathrm{s}_{{a_{1}} \neq a_{2}}(\cdot)$ denote similarity scores obtained from comparisons of non-mated samples with same and different soft-biometric attributes, respectively.

Additionally, beyond the general shift in the non-mated similarity scores distributions, the highest non-mated similarity scores (\ie those at the tail of the score distribution) tend to stem from comparisons of two non-mated samples with identical, rather than different soft-biometric attributes. In other words, as the similarity score increases, the probability of the contributing samples being associated with the same soft-biometric attribute also increases, 
\begin{equation}
    \label{eq:prob_high}
     \mathrm{s}_{a_{1}, {a_2}}(\cdot) \rightarrow 1 \Leftrightarrow P(a_{1} = {a_2}) \rightarrow 1
\end{equation} where $\mathrm{s}_{a_{1}, {a_2}}$ denotes the similarity score between two samples with soft-biometric attributes $a_1$ and $a_2$.

\begin{figure}[!t]
    \centering
    \subfigure[FaceNet]{\includegraphics[width=0.475\linewidth]{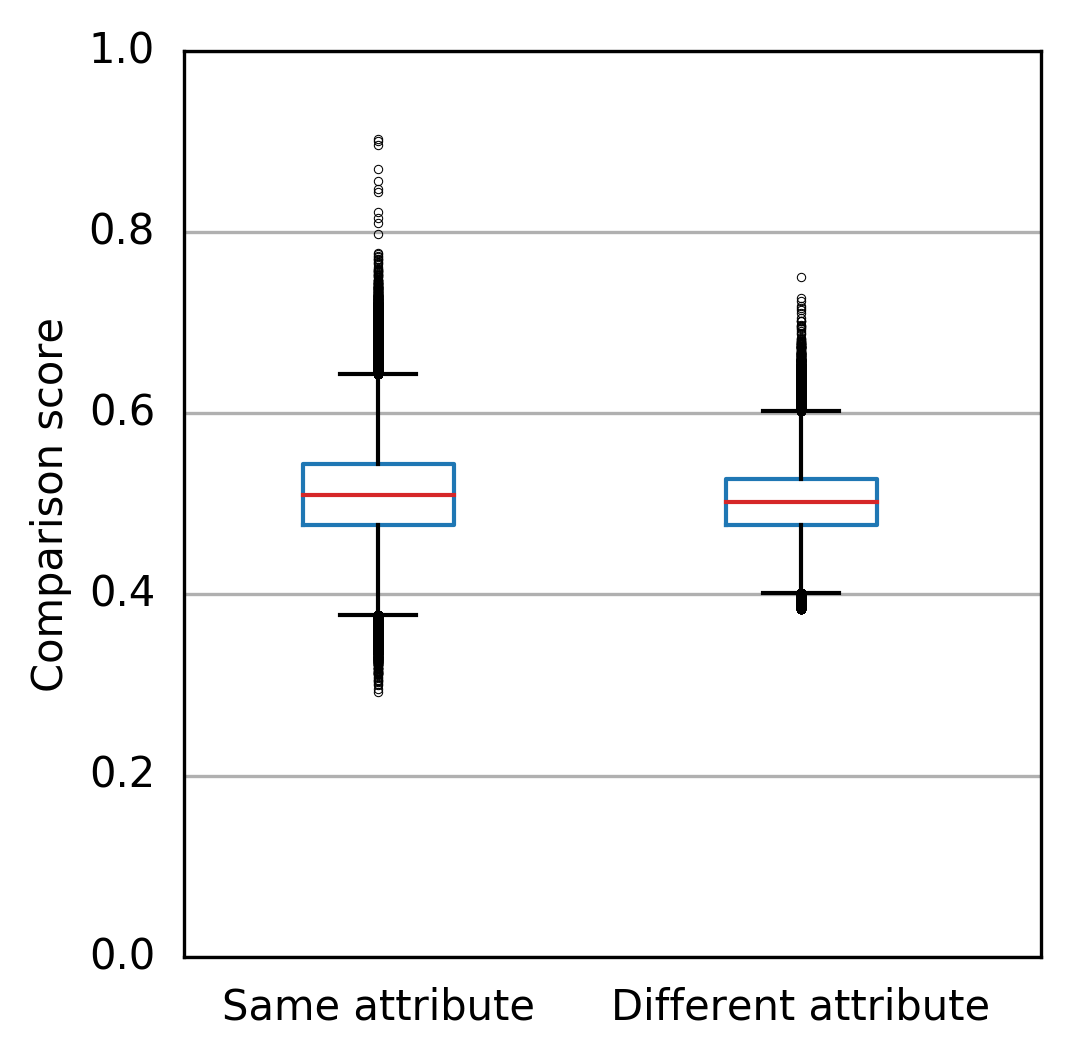}}
    \subfigure[VGGFace2]{\includegraphics[width=0.475\linewidth]{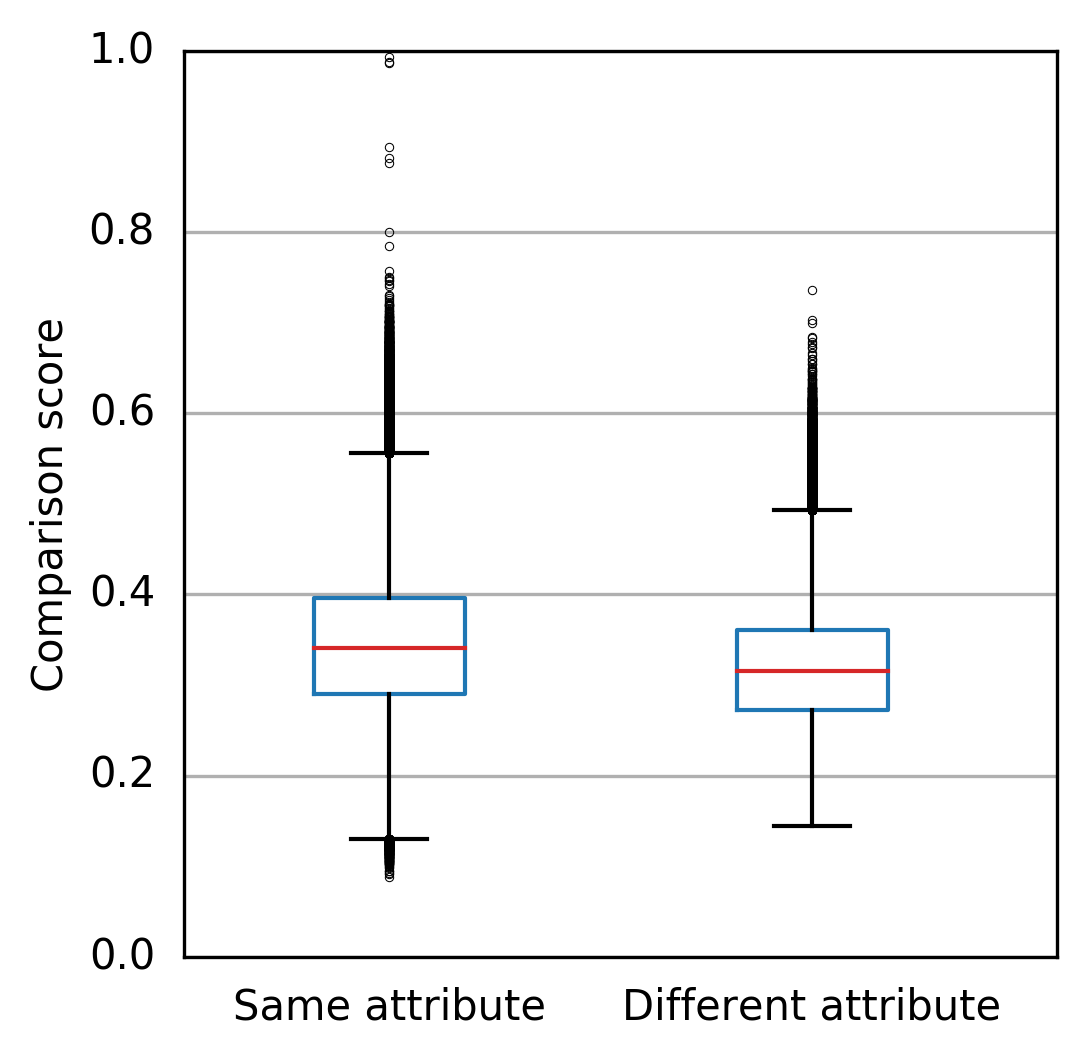}} 
    \caption{Boxplots of similarity scores for original (unprotected) non-mated comparison trials with same and different soft-biometric attributes for two face recognition systems on the LFW database. Comparison trials for the same attribute (gender) yield slightly higher similarity scores and more outliers compared to those for different attributes.}
    \label{fig:boxplots-background}
\end{figure}

Figure \ref{fig:boxplots-background} illustrates the above two propositions empirically. It can be seen, that the body of the boxplot for the ``same attribute'' similarity scores is shifted towards higher similarity scores; furthermore, its whisker and outliers are likewise shifted \wrt the boxplot for the ``different attribute'' similarity scores.

The general goal of the privacy-enhancing methods is to maintain the biometric performance and to simultaneously make infeasible inferring the soft-biometric attributes of the protected template. In other words, it is assumed that the methods retain sufficient identity information, while the information about the soft-biometric attributes is somehow disentangled/removed, \eg in  \cite{Terhorst-NoiseTransformation-2019,Morales-SensitiveNets-2020,Bortolato-FeatureDisentanglement-2020,Terhorst-SoftbiometricEnhancingPE-MIU-2020}. Intuitively, such a process appears challenging.  It would be surprising if this was possible, \ie that not even the slightest overlap between identity and \eg gender or ethnicity information existed. Thus far, this assertion has neither been theoretically proven nor rigorously tested empirically. While the privacy-enhancing methods may change the feature space to be no longer  separable (\ie prevent classification by \eg SVMs), this does not necessarily guarantee security from other types of attacks, \eg as described below.

In order to reach a decision based on a computed similarity score, biometric systems typically operate using a fixed decision threshold. Let $t$ denote such a decision threshold; if $\mathrm{s}(\cdot) > t$, the compared samples are deemed to be mated by the system. In case the samples are actually non-mated, this means a false match. %A biometric system operates perfectly iff $\forall \mathrm{s}(mated) > t \land \forall \mathrm{s}(non{-}mated) < t$. However, in practice this is not the case, \ie $\exists \mathrm{s}(mated) < t \lor \exists \mathrm{s}(non{-}mated) > t$. 
Although the feature representation and/or the comparator may operate completely different in the protected and unprotected domain, the basic principles regarding similarity scores and decision threshold remain unchanged. Hence, if biometric performance is to be maintained by the privacy-enhancing method, then the relations,

\begin{equation}
    \label{eq:threshold_prob}
        P(\mathrm{s_{unprotected}}(\cdot) > t) \equiv P(\mathrm{s_{protected}}(\cdot) > t)
\end{equation} must hold, where $\mathrm{s_{unprotected}}(\cdot)$ and $\mathrm{s_{protected}}(\cdot)$ denote similarity scores of an original and a privacy-enhancing face recognition system, respectively.

To satisfy these relations, the mapping performed by the privacy-enhancing method must be done in such a way, that sample pairs which would have achieved a high similarity score in the unprotected domain also do so in the protected domain. Due to the nearly inevitable overlap between the mated and non-mated score distributions, this implies that some non-mated sample pairs will be clustered closely together in the latent space generated by the privacy-enhancement method (\ie at least those scoring above $t$ corresponding to a certain false-match rate), \ie \textit{equation \ref{eq:prob_high} likely holds true also in the protected domain}, thus opening an attack vector. 

Bringing together the above points enables an attack aimed at inferring the soft-biometric attributes of templates protected by the aforementioned privacy-enhancing methods. The prerequisites for the attack are modest:

\begin{enumerate}
    \item The attacker intercepts a privacy-enhanced template.
    \item The attacker knows which algorithm was used to protect the template and can operate it as a black-box to generate new templates from own data.
    \item The attacker possesses or can synthesize a dataset of arbitrary facial image, with approximate balanced distribution of the target attribute.
\end{enumerate}

The attack, described in detail in the next subsection, takes advantage of the demographic differentials exhibited by most facial recognition systems, the imperfect separation between mated and non-mated distributions in the vast majority of biometric recognition systems, and other circumstances which prevent the privacy-enhancing methods to fully disentangle identity and soft-biometric information.

\subsection{Attack execution}
\label{sec:attack-analysis}
In the first step of the attack, an intercepted template is compared against the attacker's own database of privacy-enhanced facial templates. Let $N$ represent the number of samples in the attacker's database. Further, let $A = [a_{1} \dots a_{k}]$ represent the list of distinct soft-biometric attributes (\eg male and female for gender) in the attacker's database, and $k$ the count thereof. Thus, a list of $S = [s_{1} \dots s_{N}]$ similarity scores is created; furthermore, a list of same length containing the soft-biometric attributes of the samples from attacker's database is maintained.

Instead of considering the entire list of scores, only a subset of highest similarity scores is considered. Depending on the selected analysis method (described further below), the attacker selects \textit{one of the following}:

\begin{enumerate}
    \item A single list, denoted $S_{n}$, representing similarity scores taken from $S_{N}$, sorted in descending order, and cut-off after $n$ first entries.
    \item $k$ lists $S_{a_n}$, each representing similarity scores taken from $S_{N}$ for each specific attribute $a$ present in the attacker's database. The lists are sorted by similarity score in descending order, and cut-off after $n$ first entries.
\end{enumerate}

%As described in subsection \ref{sec:attack-background}, based on the behaviour of the similarity scores or attributes in the above list(s), a prediction of the soft-biometric attribute of the intercepted template is expected to be possible.
In the analysis step, the attacker applies simple algorithms or calculations to quantify the aforementioned behaviours and predict a soft-biometric attribute from a privacy-enhanced template. Let $c(a)$ represent a function which computes a loosely defined ``strength of evidence'' or a probability (not in a strict mathematical sense) of the intercepted template having a given soft-biometric attribute $a$. Further, let $C_{attack\_type} =  [c(a_{1}) \dots c(a_{k})]$ represent a list containing such probabilities for all ($k$) considered soft-biometric attributes for a given attack type:

\begin{LaTeXdescription}
\item[Majority vote] $C_{vote}$ contains the count for all $k$ possible attributes.
\item[Averaging] $C_{av}$ is a list of averages for all $k$ possible attributes.

\item[Weighted averaging] $C_{av\_lin}$ and $C_{av\_log}$  contain average similarity scores which are linearly and logarithmically weighted, respectively. Weights are assigned according to their position $i$ in the list of $n$ highest scores. Precisely, the linear weight $1-\nicefrac{i}{n+1}$ and the logarithmic weight $-\log \nicefrac{i}{n+1}$ are applied.%. 
\end{LaTeXdescription}

To reach a decision denoted $P(x)$ (\ie to predict the unknown attribute $x$ of the intercepted template), the maximum value for the chosen attack type is found, \ie $p = \underset{x}{\mathrm{argmax}}\:C_{attack\_type}$.  Finally, the corresponding soft-biometric attribute is selected accordingly, \ie $P(x) = a_{p}$.

\section{Experimental Setup}
\label{sec:experimental}
This section describes the setup of the conducted experiments. Specifically, subsection \ref{sec:experimental-setup} describes the experimental protocol, the used datasets are summarised in subsection \ref{sec:datasets}, while subsection \ref{sec:metrics} outlines the metrics used in the evaluations.

\subsection{Choice of Algorithms and Protocol}
\label{sec:experimental-setup}

Two soft–biometric privacy enhancement approaches, \ie PFRNet and PE-MIU were selected. PFRNet and PE-MIU are based on a model \cite{Cao-vggface2-2018} trained on VGGFace2 database (hereafter referred to as VGGFace2) and the FaceNet \cite{Schroff-facenet-2015} face recognition system. Accordingly, these face recognition systems are used in experiments representing the original unprotected systems. While these face recognition system may not represent the current state-of-the-art, these are used to reproduce the results reported in previous works. Additionally, the effect of broad homogeneity has recently also been confirmed for various state-of-the-art systems \cite{howard2020quantifying}.

The selection of the algorithm is based on several observations. Firstly, it is noteworthy that these methods are publicly available, \ie the experiments in this work are reproducible. Secondly, like the chosen methods, most soft–biometric privacy enhancement approaches are designed to conceal gender information, see Table~\ref{tab:overview-privacy-enhancing-solutions}. In fact, it is worth noting that there are hardly any available implementations of soft–biometric privacy enhancement methods protecting attributes other than gender. Thirdly, the two methods represent conceptually different  soft-biometric privacy enhancement approaches, \ie applied on representation level (PFRNet) and inference level (PE-MIU). Fourthly, these approaches achieved a promising trade-off between soft-biometric privacy protection and biometric performance over challenging databases such as Adience~\cite{Eidinger-Adience-2014} and LFW~\cite{Huang-wild-2007}. Although other methods do exist in the literature, they were either superseded by the aforementioned methods or their authors were not able to provide the generated templates and/or the code/models for generating them.

The evaluation consists of following parts, organised accordingly in section \ref{sec:experimental-results}:

\begin{LaTeXdescription}
    \item[Performance analysis] in a baseline evaluation, the biometric performance and gender prediction accuracy are computed using the original (unprotected) and privacy-enhanced (protected) templates, similar to the protocol described in the respective publications ~\cite{Bortolato-FeatureDisentanglement-2020,Terhorst-SoftbiometricEnhancingPE-MIU-2020}.
    \item[Vulnerability analysis] the attacks described in section \ref{sec:attacks} are carried out and their efficacy is evaluated.

\end{LaTeXdescription}
	
\subsection{Datasets}
\label{sec:datasets}
The experiments were conducted using the facial image databases with soft-biometric attribute annotations and face recognition models used by the authors of each of the considered soft–biometric privacy-enhancement approach, see Table~\ref{tab:overview-privacy-enhancing-solutions}. The privacy-enhanced templates generated by PFRNet for each dataset were provided directly by their authors~\cite{Bortolato-FeatureDisentanglement-2020}. The method was trained and applied on disjoint subsets of the CelebA database. For PE-MIU, the templates were generated using the publicly available PE-MIU software\footnote{https://github.com/pterhoer/PrivacyPreservingFaceRecognition}. This method does not require any training. The underlying face recognition models VGGFace2 and FaceNet are trained with the VGGFace2 and MS-Celeb-1M databases.

\begin{table}
	\centering
	\caption{Overview of the analysed soft–biometric privacy enhancement approaches.}
	\label{tab:overview-privacy-enhancing-solutions}
    		  \begin{adjustbox}{max width=\linewidth}
 \begin{tabular}{c c c c c} \toprule
		    \multirow{2}{*}{\textbf{Approach}} &  \textbf{Recognition}  & \textbf{Protected} &  \multirow{2}{*}{\textbf{Training}} &  \multirow{2}{*}{\textbf{Test}}   \\
		    & \textbf{Model} & \textbf{Attribute} & & \\\midrule
		   
		   \multirow{3}{*}{PFRNet \cite{Bortolato-FeatureDisentanglement-2020}} &     \multirow{3}{*}{VGGFace2~\cite{Cao-vggface2-2018}} & \multirow{3}{*}{Gender} & \multirow{3}{*}{CelebA~\cite{Liu-CelebA-2015}} & Adience~\cite{Eidinger-Adience-2014} \\
		   
		   										&       								&																		& & LFW~\cite{Huang-wild-2007} 			\\
		   										
		   										& 										&							& &										 	CelebA~\cite{Liu-CelebA-2015}					 \\ \midrule
			
			\multirow{3}{*}{PE-MIU \cite{Terhorst-SoftbiometricEnhancingPE-MIU-2020}} &   \multirow{3}{*}{FaceNet~\cite{Schroff-facenet-2015}} & \multirow{3}{*}{Gender} & \multirow{3}{*}{\textit{none}} &  Adience~\cite{Eidinger-Adience-2014}      \\
					   							 &       								 &																		& & LFW~\cite{Huang-wild-2007}					 \\
		   										
		   										& 										&																		& &	ColorFeret~\cite{Phillips-FERET-2000}	  			\\ \bottomrule
		 
    \end{tabular} 	 
	\end{adjustbox}
\end{table}

\begin{figure}[!t]
	\centering
    \includegraphics[width=0.95\linewidth]{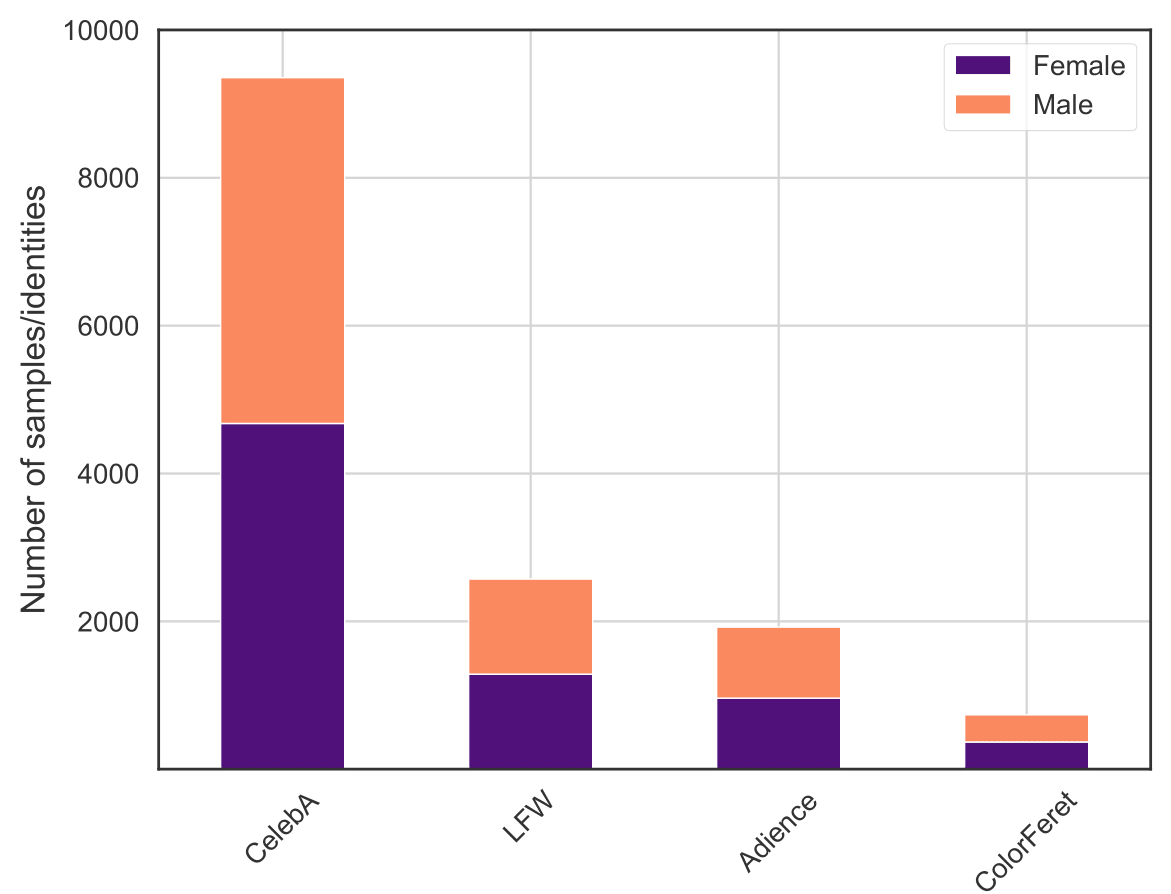}
    \caption{Gender-balanced attacker databases sorted by size.}
	\label{fig:distribution-gender}
\end{figure}

To simulate an attacker possessing their own dataset, subsets of said databases were created by selecting one sample (with highest quality) per identity. These subsets are then balanced \wrt the protected soft-biometric attribute (\ie gender) resulting in an approximate equal number of male and female subjects in the database. This is done to avoid a higher false match probability for one of the genders.
In order to avoid duplicate identities in cross-database evaluations, high similarity scores obtained from cross-database comparisons were analysed. To that end, face embeddings were extracted using a face recognition system, \ie FaceNet~\cite{Schroff-facenet-2015}, and cosine distance was used for comparison. Then, potential duplicate identities were identified by visual inspection. As a result, cross-database evaluations (\eg CelebA against LFW) containing duplicated identities were removed from our evaluations.  Figure~\ref{fig:distribution-gender} depicts an overview of the number of identities and the distribution of the gender attribute used for each dataset. Note that numerous subjects/samples have to be removed in order to obtain gender-balanced attacker databases. In the cross-database evaluations, the dataset possessed by the attacker and the dataset from which the targeted privacy-enhanced templates stem from are always different (\eg attacker is in possession of FERET database, while the target stems from LFW database).

Scenarios in which the identity of an attacked privacy-enhanced template is contained in the training database of the face recognition model or soft-biometric privacy enhancement method represent a clear disadvantage for the attacker. On the one hand, if an image from the attacker's database has been seen by the recognition model during training, it is expected that it is more easily separable from other identities and, hence, it is less likely to produce a high similarity score. On the other hand, in case an image from the attacker's database has been seen by the privacy enhancing technique during training, \ie PFRNet, it can be assumed that gender information will be suppressed more effectively for this identity. Thus, this identity has less chance to produce a high similarity score with an attacked template of the same gender.
	
\subsection{Metrics} 
\label{sec:metrics}
The experimental evaluation is conducted according to ISO/IEC 19795-1~\cite{ISO-IEC-19795-1-Framework-210216} standard methods. The standard and additional metrics used in the experimental evaluation are as follows:

\begin{LaTeXdescription}
    \item[Biometric performance] the False Non-Match Rate (FNMR) and False Match Rate (FMR) denote the proportion of falsely classified mated and non-mated attempts in a biometric verification scenario, respectively. Additionally, the equal error rate (EER), which is the point where FMR and FNMR are equal, is reported.
    \item[Attack success rate] percentage of samples correctly classified in terms of soft-biometric attribute by an attack. This rate can also be seen as gender prediction accuracy.
\end{LaTeXdescription}

\section{Results}
\label{sec:experimental-results}
In this section, subsection~\ref{sec:performance-analysis} presents an performance analysis of the used soft-biometric privacy-enhancing approaches. Subsequently, a vulnerability analysis of said methods to the proposed attack is conducted in subsection~\ref{sec:vulnerability-analysis}.

\begin{table}[!t]
	\scriptsize
	\centering
	\caption{Biometric performance for original (unprotected) and privacy-enhanced (protected) systems (in \%).}
	\label{tab:biometric-performance}
    	\begin{adjustbox}{max width=\linewidth}
	\begin{tabular}{cc ccc ccc } 
	\toprule
	
	              \multirow{2}{*}{\textbf{Method}} &   \multirow{2}{*}{\textbf{Dataset}}              &\multicolumn{3}{c}{\textbf{Original}}&\multicolumn{3}{c}{\textbf{Privacy-enhanced}}\\ \cmidrule(r){3-5} \cmidrule(l){6-8}
	
   &&                        \textbf{EER}&\textbf{FMR}&\textbf{FNMR}&           \textbf{EER}&\textbf{FMR}&\textbf{FNMR}\\ \midrule
   
     \multirow{9}{*}{PFRNet}&\multirow{3}{*}{LFW}&              \multirow{3}{*}{0.80}&0.001&3.562&\multirow{3}{*}{1.38}&0.001&7.098    \\
                  &                &                           &                     0.01&0.665 &               &0.01 &1.762    \\
                  &                &                           &                     0.1 &0.066 &               &0.1  &0.231    \\\cmidrule{2-8}
                  
                  &\multirow{3}{*}{Adience}&                  \multirow{3}{*}{6.70}&0.001&74.445 &        \multirow{3}{*}{6.72}&0.001&80.191   \\
                  &                &                           &                     0.01&40.022&           &0.01  &43.501   \\
                  &                &                           &                     0.1 &4.332 &            &0.1  &4.632    \\ \cmidrule{2-8}

                  &\multirow{3}{*}{CelebA}&\multirow{3}{*}{6.47}&                    0.001&30.786    &\multirow{3}{*}{9.37}               &0.001&34.326        \\
                  &                &                           &                     0.01&15.533   &               &0.01  &20.108    \\
                  &                &                           &                     0.1 &4.962   &               &0.1  &9.034    \\ \midrule

     \multirow{9}{*}{PE-MIU}&\multirow{3}{*}{LFW}&              \multirow{3}{*}{0.55}&0.001&2.116&\multirow{3}{*}{0.64}&0.001&2.234    \\
                  &                &                           &                     0.01&0.347&               &0.01 &0.512    \\
                  &                &                           &                     0.1 &0.049&               &0.1  &0.165    \\ \cmidrule{2-8}

                  &\multirow{3}{*}{Adience}&                  \multirow{3}{*}{4.72}&0.001&63.675&        \multirow{3}{*}{4.72}&0.001&63.675    \\
                  &                &                           &                     0.01&22.640&           &0.01  &22.638   \\
                  &                &                           &                     0.1 &2.057 &            &0.1  &2.347  \\ \cmidrule{2-8}

                  &\multirow{3}{*}{ColorFeret}&\multirow{3}{*}{2.16}&                0.001&16.721   &\multirow{3}{*}{2.70}&0.001&16.724         \\
                  &                &                           &                     0.01&4.083    &               &0.01  &4.613   \\
                  &                &                           &                     0.1 &0.419   &               &0.1  &1.246   \\
	
	\bottomrule
			\end{tabular}
	\end{adjustbox}
\end{table}
\begin{table*}[!t]
	\scriptsize
	\centering
	\caption{Gender prediction performance of basic machine learning-based classifiers on orginal (unprotected) and privacy-enhanced (protected) templates in cross-database scenarios (in \%).}
	\label{tab:gender-prediction}
    	\begin{adjustbox}{max width=\linewidth}
		\begin{tabular}{ccc c ccc c ccc } 
		\toprule
		
		 \multirow{3}{*}{\textbf{Method}}&\multirow{3}{*}{\textbf{Training}}   &\multirow{3}{*}{\textbf{Testing}}  & \multicolumn{4}{c}{\textbf{Original}} & \multicolumn{4}{c}{\textbf{Privacy-enhanced}} \\ 
		 &  &  &    &\multicolumn{3}{c}{\textbf{SVM}} & &\multicolumn{3}{c}{\textbf{SVM}}\\
		 \cmidrule{5-7}
		 \cmidrule{9-11}
		
		  &    &  & \textbf{kNN} &\textbf{Poly}&\textbf{RBF}&\textbf{Sigmoid}&\textbf{kNN}  &\textbf{Poly}&\textbf{RBF}&\textbf{Sigmoid}\\ \midrule
		 
		  \multirow{4}{*}{PFRNet}              &LFW                  &Adience                           &80.80    &\textbf{91.00}&90.95&87.36&69.82&79.60 &79.70&\textbf{62.64} \\   \cmidrule{2-11}    
		  
		                                        &\multirow{2}{*}{Adience} &CelebA                         &77.40    &91.24&\textbf{94.54}&87.84 &63.16  &59.33&58.42&\textbf{51.19}\\
		                                        &                         &LFW                            &83.36    &94.56&\textbf{96.93}&87.91 &73.79  &76.39&75.42&\textbf{67.80}\\ \cmidrule{2-11}
		                                         
		                                        &CelebA                   &Adience                        &80.75    &89.65&\textbf{91.42}&86.73 &73.20 &83.71&86.00&\textbf{70.00}\\\midrule

		  \multirow{6}{*}{PE-MIU}                &\multirow{2}{*}{LFW}    &Adience                         &90.42    &\textbf{91.31}&87.04&60.21 &58.55 &56.20 &64.06&\textbf{50.00}\\
		                                         &                        &ColorFeret                      &95.38    &89.13&\textbf{96.20}&68.34 &61.41 &\textbf{55.43} &70.92&59.78\\ \cmidrule{2-11}
		                                         &\multirow{2}{*}{Adience}&LFW                             &95.18    &89.81&\textbf{96.73}&72.86 &59.18 &57.93 &63.06&\textbf{54.74}\\ 
		                                         &                        &ColorFeret                      &\textbf{88.99}    &84.24&88.18&80.98 &62.09 &60.46 &61.68&\textbf{57.20}\\ \cmidrule{2-11}
		                                         
		                                         &\multirow{2}{*}{ColorFeret}&LFW                          &93.51    &84.56&\textbf{98.21}&75.00 &57.62 &50.27 &\textbf{50.00}&51.63\\
		                                         &                           &Adience                      &85.17    &80.84&\textbf{89.54}&77.39 &59.22 &\textbf{50.00} &50.36&51.88\\

\bottomrule

		\end{tabular}
	\end{adjustbox}
\end{table*}

\subsection{Performance analysis}
\label{sec:performance-analysis}
As a first step, the biometric performance of the unprotected systems, \ie original system, is estimated and compared against that of the corresponding privacy-enhanced systems. In Table~\ref{tab:biometric-performance}, the face verification performance is reported on different databases for each method. PFRNet and PE-MIU have both been applied on LFW and Adience. In addition PFRNet has been applied to CelebA and PE-MIU to ColorFeret, respectively. Based on the obtained results we can observe that the verification performance on privacy-enhanced system is slightly degraded compared to the original system. This confirms that privacy enhancement defines a trade-off between identity information and suppression of privacy-sensitive attributes, as it is shown in~\cite{Terhorst-SoftbiometricEnhancingPE-MIU-2020}. It can be observed that for FMRs$\textless 0.1\%$, the PE-MIU biometric performance is up to seven times lower than the PFRNet performance over similar databases, \ie LFW and Adience. In particular, for a practical scenario (FMR$ = 0.1\%$), PE-MIU rejects only approximately 0.17\% and 2.35\% of the mated samples over these two challenging databases, respectively. Overall, both systems obtain impressive performance rates across different databases which are generally retained in their privacy-enhanced versions.

\begin{figure}[!t]
    \centering
    \subfigure[PFRNet]{\includegraphics[width=0.475\linewidth]{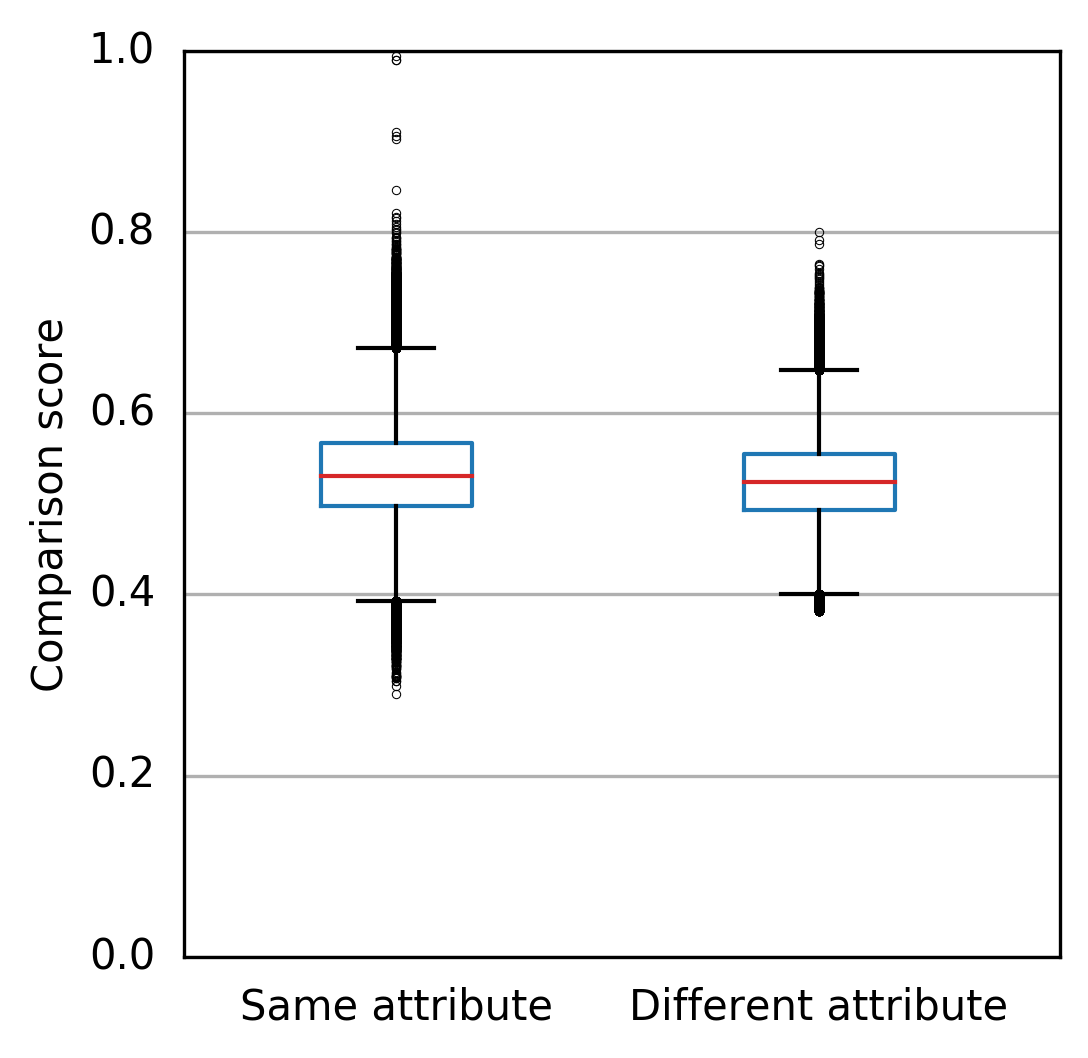}}
    \subfigure[PE-MIU]{\includegraphics[width=0.475\linewidth]{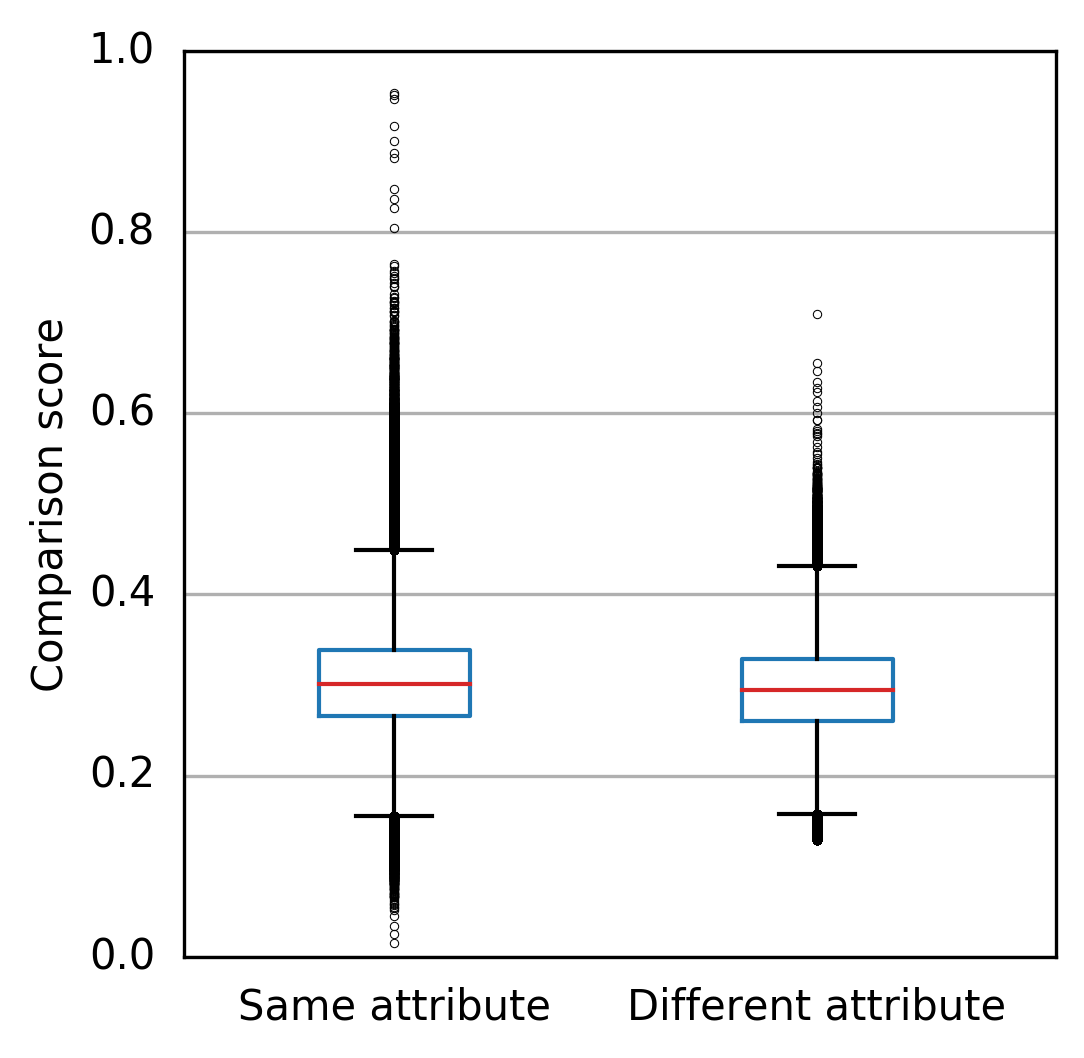}} 
    \caption{Boxplots of similarity scores for non-mated comparison trials of privacy-enhanced templates with same and different soft-biometric attributes for both algorithms on the LFW database. Comparison trials for the same attribute (gender) yield slightly higher similarity scores and more outliers compared to those for different attributes.}
    \label{fig:boxplots-attack}
\end{figure}

In the second experiment, the gender prediction performance of both both approaches, \ie PRF-Net and PE-MIU, is explored. To that end, machine learning-based gender classifiers are trained on original face embeddings and privacy-enhanced templates obtained by both approaches in cross-database scenarios, \eg training on LFW and gender prediction on Adience, where the number of subjects for each gender attribute is also balanced (see Figure~\ref{fig:distribution-gender}). In Table~\ref{tab:gender-prediction}, the gender prediction performance is reported for different classic classifiers, \eg kNN and SVM. Here, SVM is employed by training different kernels (Poly, RBF, and Sigmoid). Note that hyper-parameters of both classifiers were set to basic configurations without optimisation.

 \begin{figure}[!t]
% 	\centering
\subfigure[Original -- VGGFace2]{
\includegraphics[width=0.48\linewidth,trim=22 22 22 22,clip]{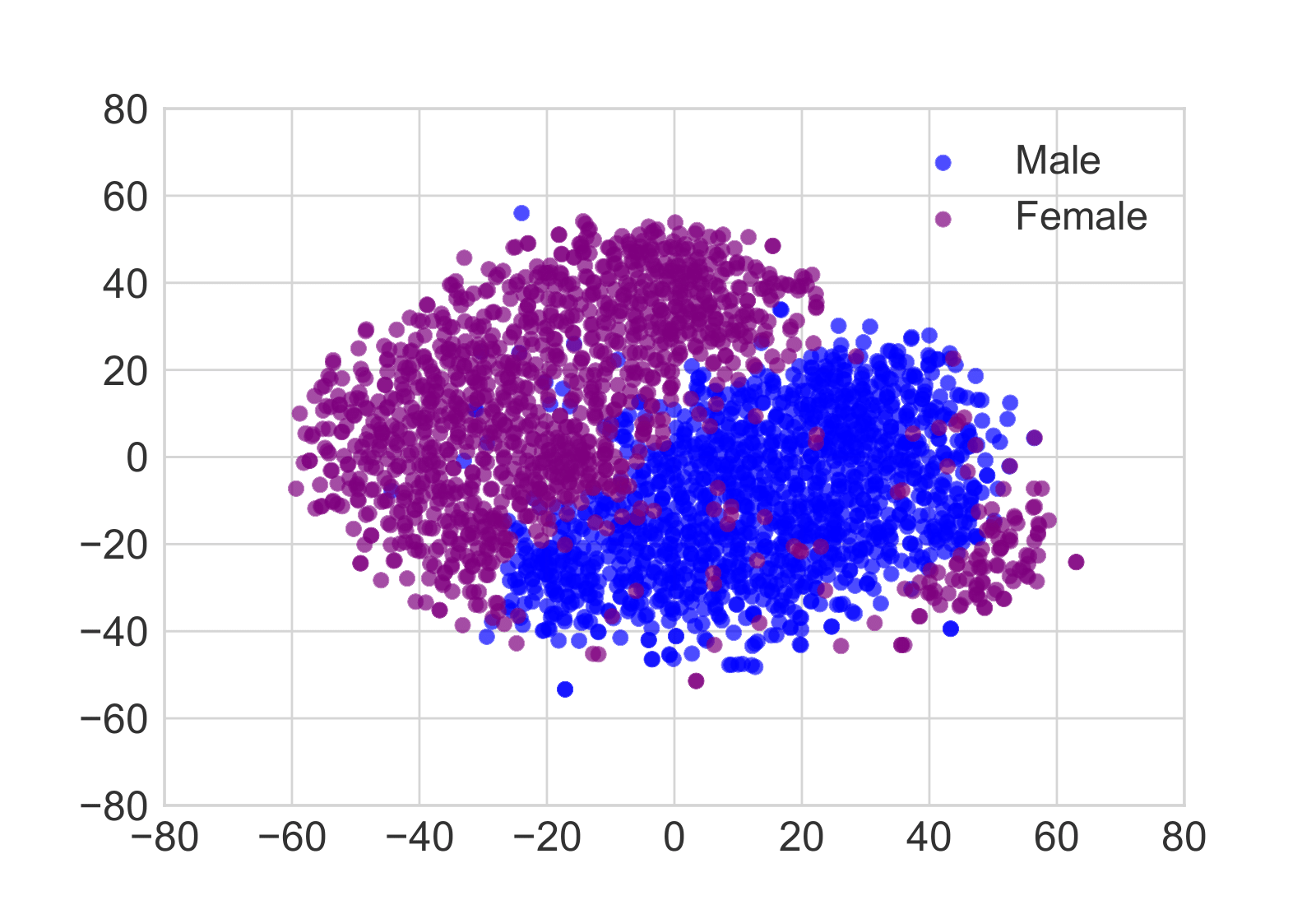}} 
\subfigure[Privacy-enhanced -- PRF-Net]{
\includegraphics[width=0.48\linewidth,trim=22 22 22 22,clip]{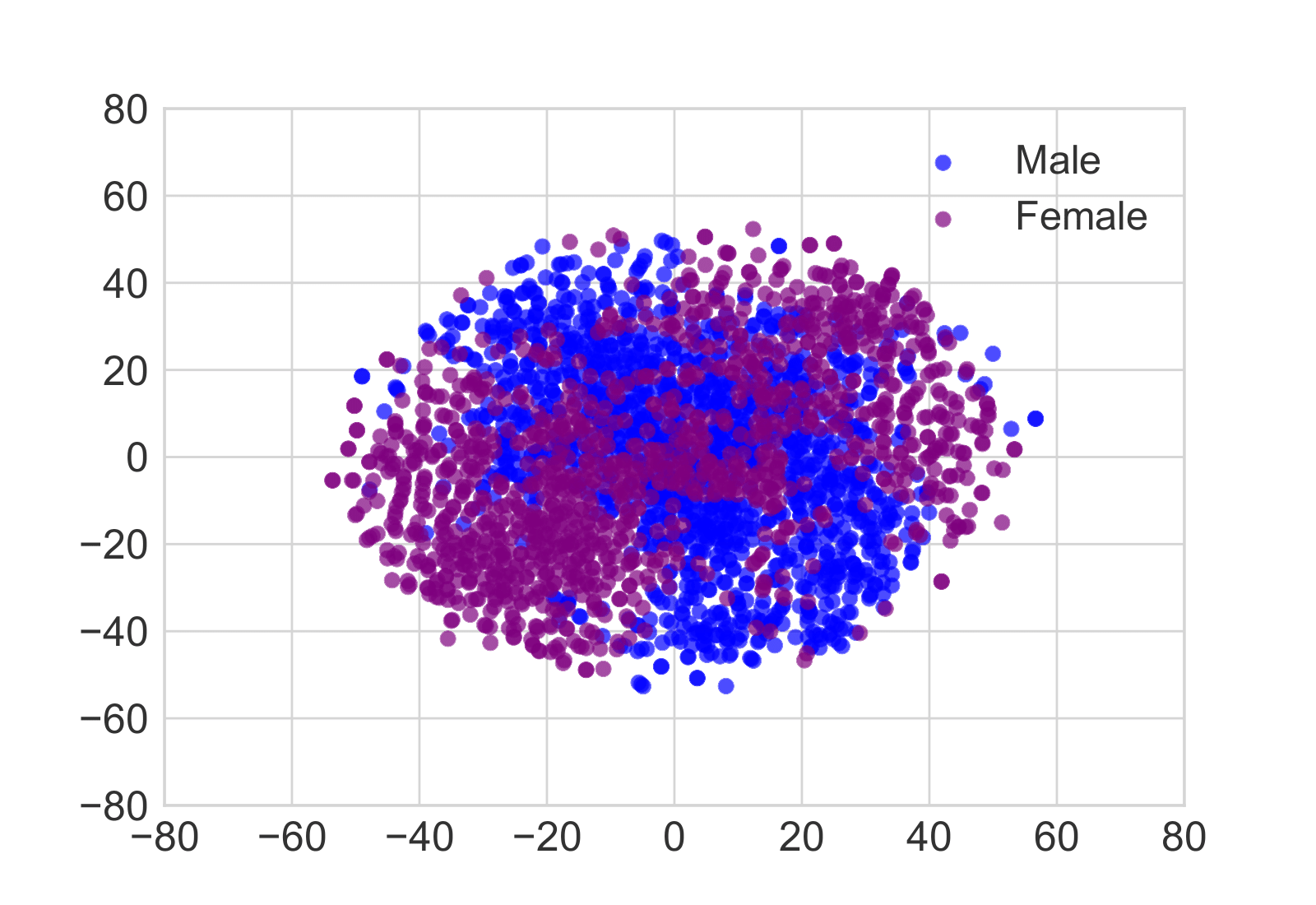}} 
\subfigure[Original -- FaceNet]{
    \includegraphics[width=0.48\linewidth,trim=22 22 22 22,clip]{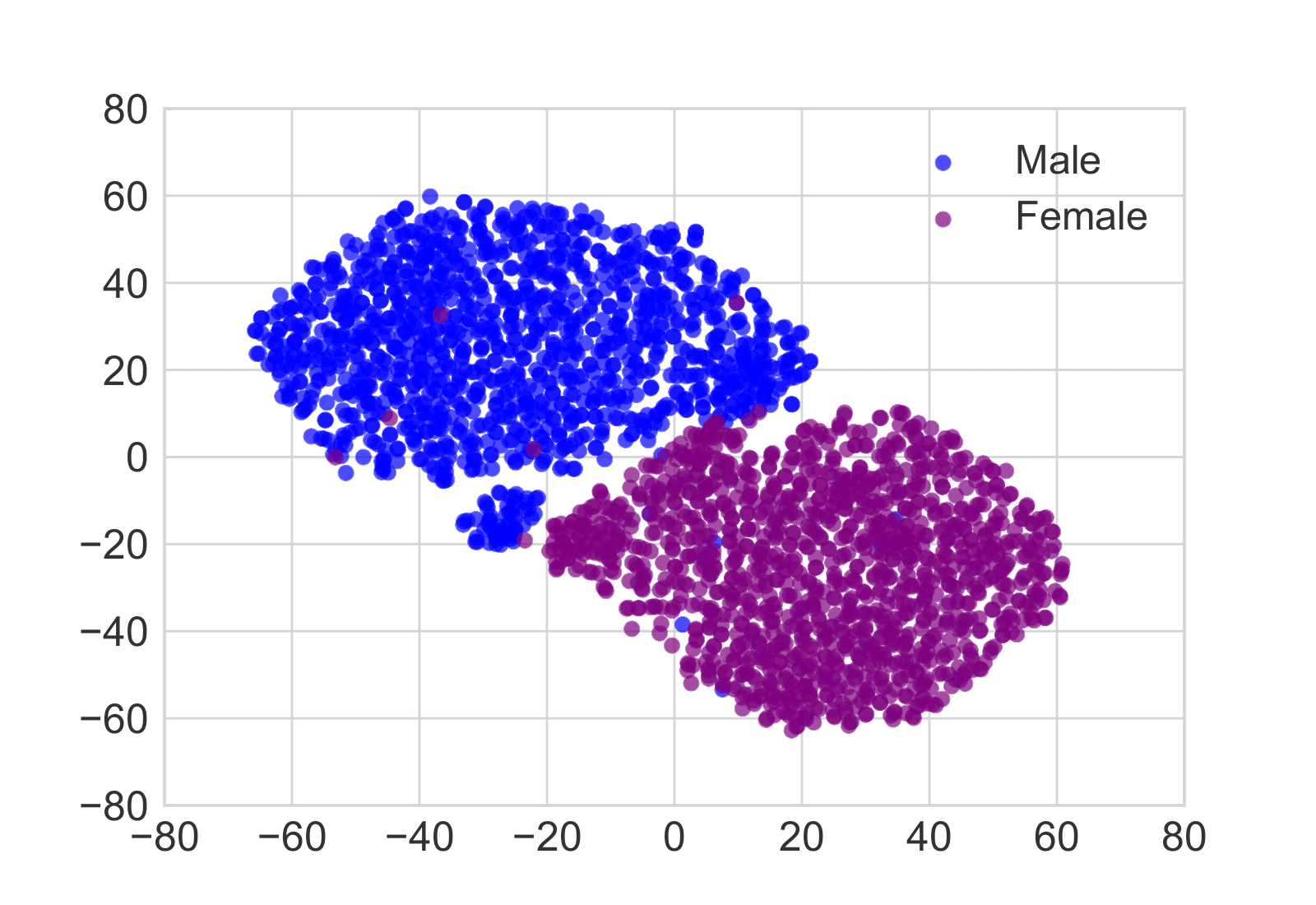}} 
\subfigure[Privacy-enhanced -- PE-MIU]{
    \includegraphics[width=0.48\linewidth,trim=22 22 22 22,clip]{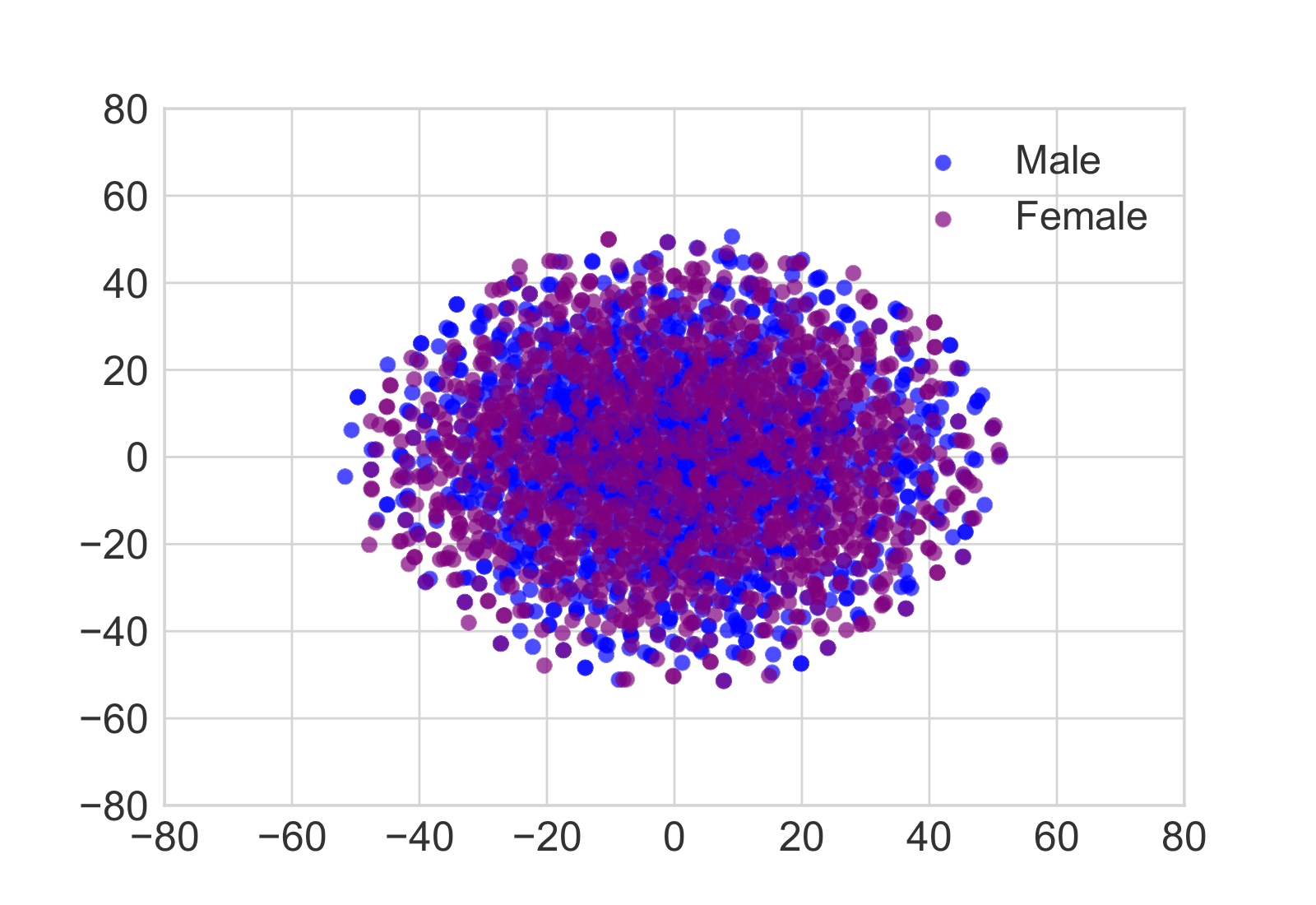}}
    \caption{Visualization of original (unproteced) and privacy-enhanced (protected) face representations over the LFW database using t-SNE.}
	\label{fig:t-sne-visualization}
\end{figure} 

\begin{figure*}[!t]
\includegraphics[width=\linewidth]{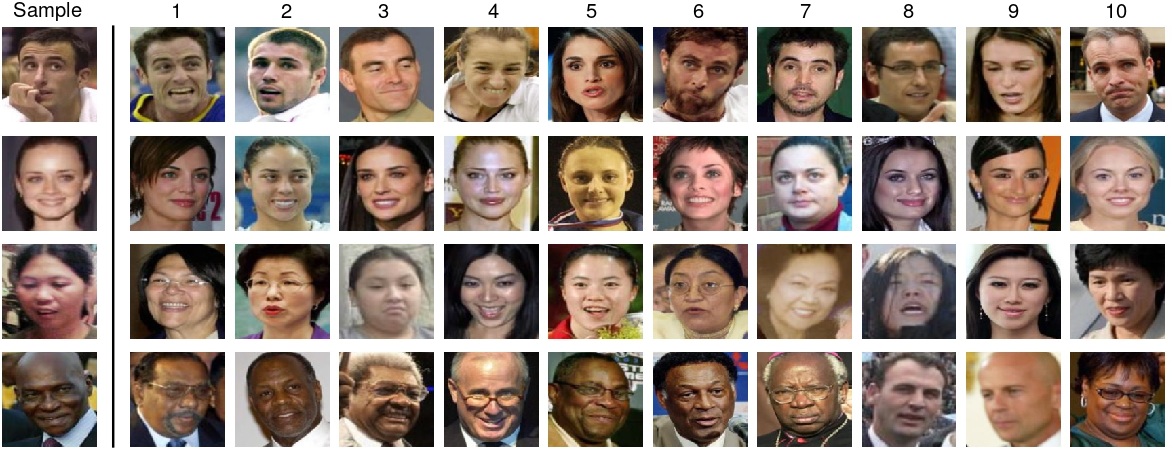}
\caption{Ranked examples of samples that reach a high similarity score in a non-mated comparison of privacy-enhanced templates of PFRNet (first and second row) and PE-MIU (third and fourth row); images taken from the LFW database.}
    \label{fig:best-scores}
\end{figure*}

A significant degradation of the gender prediction performance is observable for privacy-enhanced templates compared to unprotected templates. Lowest average gender prediction accuracy of 52.37\% is obtained by PE-MIU, in contrast to 65.22\% for PFRNet, over similar cross-database scenarios (\ie training on LFW and Adience to predict on Adience and LFW respectively). These results indicate that machine learning-based classifiers are not able to reliably predict gender from privacy-enhanced templates. This is further supported by looking at visualisations obtained by dimensionality reduction tools.  Examples using t-distributed stochastic neighbour embeddings (t-SNE)~\cite{Van-VisualizingT-SNE-2008} are depicted in Figure~\ref{fig:t-sne-visualization}. It can be observed that in their original embeddings, faces are clustered with respect to gender, which is not the case for the privacy-enhanced templates. At this point, it is important to repeat that such observations are the basis for reporting high level of soft-biometric privacy in some published works, \eg in \cite{Terhorst-NoiseTransformation-2019,Morales-SensitiveNets-2020,Bortolato-FeatureDisentanglement-2020,Terhorst-SoftbiometricEnhancingPE-MIU-2020}.

\subsection{Vulnerability analysis}
\label{sec:vulnerability-analysis}

In a first experiment, it is analysed whether the propositions about the properties of privacy-enhanced templates hold. Figure~\ref{fig:boxplots-attack} depicts examples of similarity scores for non-mated comparison trails of privacy-enhanced templates with same and different soft-biometric attributes for both used methods (analogous to Figure~\ref{fig:boxplots-background}). Like in the original unprotected systems, ``same attribute''  boxplots are shifted towards higher similarity score. In addition, facial image pairs which produce high similarity scores when comparing their corresponding privacy-enhanced templates have been visually inspected. Examples of  samples and top-ranked samples that obtain high similarity scores in non-mated comparison of privacy-enhanced template are depicted in Figure~\ref{fig:best-scores}. It can be seen that with high probability the gender of top-ranked samples is the same as that of the leftmost sample. This suggests that the effect of broad homogeneity still exists in the protected domain.   

\begin{table*}[!t]
	\scriptsize
	\centering
	\caption{Attack success rates of the attack employing the majority voting strategy (in \%).}
	\label{tab:mojority-voting}
	\resizebox{0.7\linewidth}{!}{\begin{adjustbox}{max width=\linewidth}
	\begin{tabular}{ccc cccccc}
	\toprule
	
		\multirow{2}{*}{\textbf{Method}}&\multirow{2}{*}{\textbf{Attacker}}&\multirow{2}{*}{\textbf{Target}}   & \multicolumn{6}{c}{\textbf{Attack success rate}}  \\ 
	
	                   &                                   &                  &$n=1$&$n=5$&$n=11$&$n=51$&$n=101$&$n=201$ \\    \midrule
	                   \multirow{4}{*}{PFRNet}         &LFW                  &Adience   &77.14&82.15&\textbf{82.50}&80.17&78.62&76.13 \\ \cmidrule{2-9}

	                   &\multirow{2}{*}{Adience} &CelebA    &76.69&79.71&\textbf{80.91}&80.12&79.08&77.63    \\

	                   &                         &LFW     &75.29&78.30&\textbf{78.36}&76.48&74.56&70.14\\ \cmidrule{2-9}

	                   &CelebA                   &Adience &70.56&\textbf{71.83}&71.06&66.32&63.75&60.35 
	                                                       
	                                                       \\ \midrule
	                                                       
	\multirow{6}{*}{PE-MIU}             &\multirow{2}{*}{LFW}  &Adience                &86.51&\textbf{87.99}&87.67&85.03&82.58&78.58 \\
	
	                                    &                        &ColorFeret  &83.44&84.18&\textbf{84.33}&81.53&78.73&72.74\\ \cmidrule{2-9}
	                                    
	                                    &\multirow{2}{*}{Adience}&LFW 
	                                    &86.05&87.40&\textbf{87.92}&84.59&82.88&79.96\\
	                                    
	                                    &                        &ColorFeret   &79.54&81.57&\textbf{81.73}&77.98&74.75&70.01 \\ \cmidrule{2-9}
	                                    
	                                    &\multirow{2}{*}{ColorFeret}&LFW   &91.03&92.39&\textbf{92.80}&91.98&89.13&86.01 \\
	                                    
	                                    &                           &Adience  &84.78&\textbf{87.91}&87.77&86.96&84.92&84.38  \\ 

	\bottomrule
	\end{tabular}
	\end{adjustbox}}
\end{table*}

\begin{table*}[!t]
	\scriptsize
	\centering
	\caption{Attack success rates of the attack employing the averaging strategy (in \%).}
	\label{tab:average-scores}
    \resizebox{0.7\linewidth}{!}{% 	\begin{adjustbox}{max width=\linewidth}

	\begin{adjustbox}{max width=\linewidth}
	\begin{tabular}{ccc cccccc}
	\toprule
	
	\multirow{2}{*}{\textbf{Method}}&\multirow{2}{*}{\textbf{Attacker}}&\multirow{2}{*}{\textbf{Target}}   &  \multicolumn{6}{c}{\textbf{Attack success rate}}   \\ 
	
	                                &                                   &                  &$n=1$&$n=5$&$n=10$&$n=50$&$n=100$&$n=200$   \\    \midrule
	
	\multirow{4}{*}{PFRNet}              &LFW                  &Adience     &77.14&83.36&\textbf{83.59}&81.61&79.74&77.41\\ \cmidrule{2-9}
	                                                                        
	                   &\multirow{2}{*}{Adience} &CelebA   &76.69&80.49&\textbf{81.37}&81.32&80.70&78.77\\
	                                                      
	                   &                         &LFW     &75.29&79.45&\textbf{79.97}&77.47&75.81&72.06 \\ \cmidrule{2-9}
	                   
	                   &CelebA                   &Adience   &70.56&\textbf{72.71}&72.35&67.84&64.98&61.80\\ \midrule
	                                                       
	\multirow{6}{*}{PE-MIU}             &\multirow{2}{*}{LFW}    &Adience   &86.51&89.74&90.40&88.65&86.16&83.16   \\
	
	                                    &                        &ColorFeret  &83.44&86.82&\textbf{86.98}&84.56&81.77&78.38 \\ \cmidrule{2-9}
	                                    
	                                    &\multirow{2}{*}{Adience}&LFW     &86.05&88.60&\textbf{89.02}&87.82&86.05&83.71\\
	                                    
	                                    &                        &ColorFeret  &79.54&83.08&\textbf{83.13}&81.05&78.87&76.58\\ \cmidrule{2-9}
	                                    
	                                    &\multirow{2}{*}{ColorFeret}&LFW  &91.03&92.39&\textbf{93.89}&93.21&91.58&89.40 \\
	                                    
	                                    &                           &Adience  &84.78&87.91&87.77&\textbf{88.59}&86.96&85.73\\
	                                   
	\bottomrule
	\end{tabular}
	\end{adjustbox}}
\end{table*}

\begin{table*}[!t]
	\scriptsize
	\centering
	\caption{Attack success rates of the attack employing the linearly weighted averaging strategy (in \%).}
	\label{tab:linear-based weight}
    \resizebox{0.7\linewidth}{!}{% 	\begin{adjustbox}{max width=\linewidth}

		\begin{adjustbox}{max width=\linewidth}
	\begin{tabular}{ccc cccccc}
	\toprule
	
	\multirow{2}{*}{\textbf{Method}}&\multirow{2}{*}{\textbf{Attacker}}&\multirow{2}{*}{\textbf{Target}}   & \multicolumn{6}{c}{\textbf{Attack success rate}}  \\ 
	
	                                &                                   &                  &$n=1$&$n=5$&$n=10$&$n=50$&$n=100$&$n=200$\\    \midrule
	
	\multirow{4}{*}{PFRNet}         &LFW                  &Adience   &77.14&82.08&\textbf{83.67}&82.78&81.18&79.12 \\ \cmidrule{2-9}

	                   &\multirow{2}{*}{Adience} &CelebA    &76.69&79.81&81.48&\textbf{81.79}&81.32&80.12      \\

	                   &                         &LFW     &75.29&78.82&\textbf{79.60}&78.67&76.80&74.92 \\ \cmidrule{2-9}

	                   &CelebA                   &Adience &70.56&\textbf{72.84}&72.54&69.45&66.97&64.06 \\
	                                                       
	                                                        \midrule
	                                                       
	\multirow{6}{*}{PE-MIU}             &\multirow{2}{*}{LFW}  &Adience &86.51&89.93&90.16&\textbf{90.32}&88.37&85.73\\
	
	                                    &                        &ColorFeret  &83.44&86.20&\textbf{87.40}&85.93&84.21&81.30 \\ \cmidrule{2-9}
	                                    
	                                    &\multirow{2}{*}{Adience}&LFW  &86.05&88.44&\textbf{88.91}&88.55&87.61&85.74  \\
	                                    
	                                    &                        &ColorFeret   &79.54&82.04&\textbf{83.29}&82.72&81.10&78.81 \\ \cmidrule{2-9}
	                                    
	                                    &\multirow{2}{*}{ColorFeret}&LFW   &91.03&91.85&93.34&\textbf{93.75}&92.93&91.30 \\
	                                    
	                                    &                           &Adience  &84.78&87.50&87.91&\textbf{89.67}&88.45&87.09\\

	\bottomrule
	\end{tabular}
	\end{adjustbox}}
\end{table*}

In the second experiment, the different types of attacks are launched to derive the gender attribute from privacy-enhanced templates. It is important to note that all the attack strategies are analysed in cross-database scenarios. In the first step, the attack is applied using the majority-based voting strategy to derive gender from privacy-enhanced templates. Obtained results are summarised in Table~\ref{tab:mojority-voting} where best obtained results for each cross-database scenario are marked bold. Scenarios in which a web-collected face image database (Adience, LFW, or CelebA) are used in the attack is considered most relevant since an attacker could effortlessly access and collect such images. Employing the majority voting-based strategy, the attacker obtains the gender attribute from the $n$ odd best scores. Highest attack success rates are achieved for employing a small number of $n=11$ best scores. The average obtained attack success rates for this attack strategy lies around 85\% which is clearly above that  achieved  by machine learning-based classifiers (\cf Table~ \ref{tab:gender-prediction} on the right hand side for privacy-enhanced templates).

\begin{table*}[!t]
	\scriptsize
	\centering
	\caption{Attack success rates of the attack employing the logarithmically weighted averaging strategy (in \%).}
	\label{tab:logarithm-based weight}
    \resizebox{0.7\linewidth}{!}{% 	\begin{adjustbox}{max width=\linewidth}

	\begin{adjustbox}{max width=\linewidth}
	\begin{tabular}{ccc cccccc}
	\toprule
	
	\multirow{2}{*}{\textbf{Method}}&\multirow{2}{*}{\textbf{Attacker}}&\multirow{2}{*}{\textbf{Target}}   & \multicolumn{6}{c}{\textbf{Attack success rate}} \\
	
	                                &                                   &                  &$n=1$&$n=5$&$n=10$&$n=50$&$n=100$&$n=200$\\    \midrule
	
	\multirow{4}{*}{PFRNet}         &LFW                  &Adience  &77.14&81.18&83.24&\textbf{83.44}&82.19&81.07\\ \cmidrule{2-9}

	                   &\multirow{2}{*}{Adience} &CelebA       &76.69&79.19&80.39&\textbf{82.15}&81.32&81.11    \\

	                   &                         &LFW      &75.29&78.67&78.98&\textbf{79.45}&78.04&76.27 \\\cmidrule{2-9}

	                   &CelebA                   &Adience   &70.56&72.48&\textbf{72.89}&70.79&68.62&66.31
	                                                       \\ \midrule
	                                                       
	\multirow{6}{*}{PE-MIU}             &\multirow{2}{*}{LFW}    &Adience &86.51&89.39&90.12&\textbf{90.75}&90.05&88.34 \\
	
	                                    &                        &ColorFeret  &83.44&85.46&\textbf{87.05}&86.90&85.65&83.59 \\ \cmidrule{2-9}
	                                    
	                                    &\multirow{2}{*}{Adience}&LFW   &86.05&88.18&88.81&\textbf{88.86}&88.34&87.30    \\
	                                    
	                                    &                        &ColorFeret    &79.54&81.31&82.82&\textbf{83.08}&82.61&80.95  \\ \cmidrule{2-9}
	                                    
	                                    &\multirow{2}{*}{ColorFeret}&LFW  &91.03&91.71&92.26&\textbf{94.43}&94.02&92.93 \\
	                                    
	                                    &                           &Adience &84.78&87.50&87.91&\textbf{89.54}&88.86&88.32 \\

	\bottomrule
	\end{tabular}
	\end{adjustbox}}
\end{table*}

\begin{table*}[!t]
\scriptsize
	\centering
	\caption{Summary of the best average attack success rates across all cross-database scenarios (in \%).}
	\label{tab:summary}
	\resizebox{0.7\linewidth}{!}{

	\begin{adjustbox}{max width=\linewidth}
	\begin{tabular}{ccc cccc}
	\toprule
	
	\multirow{2}{*}{\textbf{Method}}& \multicolumn{6}{c}{\textbf{Attack success rate}}  \\ 
	
	                               &$n=1$&$n=5$&$n=10$&$n=50$&$n=100$&$n=200$\\   \midrule
	
	PFRNet    &74.92 $\pm$ 4.79&79.04 $\pm$ 7.08&79.50 $\pm$ 7.42&78.96 $\pm$ 9.06&77.54 $\pm$ 9.88&75.63 $\pm$ 12.79 \\

	PE-MIU    &85.23 $\pm$ 3.97&88.12 $\pm$ 3.28&88.65 $\pm$ 3.68&88.95 $\pm$ 3.99&88.26 $\pm$ 4.08&86.91 $\pm$ 4.38 \\

	\bottomrule
	\end{tabular}
	\end{adjustbox}
	}
\end{table*}

Table~\ref{tab:average-scores}, Table~\ref{tab:linear-based weight}, and Table~\ref{tab:logarithm-based weight} list the attack success rates for the averaging strategy. Again, best attack success rates for each cross-database experiment are marked bold. In this attack strategies $n$ best scores against male and female subjects form the attacker database are averaged and compared to obtain the gender attribute from privacy-enhanced templates. For averaging without weights, competitive attack success rates are achieved for considering $n=10$ best males and females scores. Overall, slight improvements (up to approximately two percent points) are observable when comparing the averaging strategies to the majority voting-based strategy. Further, in case scores are weighted \wrt their rank, higher values of $n\geq50$ can reveal improved attack success rates (around one percent point) compared to the simple averaging. Moreover, by weighting the scores the attack is expected to become more robust, \ie less sensitive to $n$.  

The mentioned disadvantage for the attacker (overlapping identities in the training database of the  privacy enhancement method and the attacker's database)  becomes clear for the scenario where the privacy-enhanced templates produced by PFRNet are attacked using the CelebA as attacker database. Here, the attack success chances are generally lower compared to the other evaluated scenarios.

In summary, the obtained results confirm that both analysed schemes, \ie PE-MIU and PFRNet, are highly vulnerable to the proposed attack. For a better overview, Table~\ref{tab:summary} summarises the best average attack success rates across all cross-database scenarios for different values of $n$ with a 95\% confidence interval.

\section{Discussion}
\label{sec:discussion}

This section discusses different relevant aspects of the attack. Subsection~\ref{sec:prevent} describes potential countermeasures against the attack. Alternative attack methods are briefly discussed in subsection~\ref{sec:alternative}.  The application of the proposed attack to systems based on other biometric characteristics is discussed in subsection~\ref{sec:other}. Finally, different attack models are described in subsection~\ref{sec:models}.

\subsection{Attack Prevention}\label{sec:prevent}
 
The proposed attack may be prevented by other techniques which meet the goal of protecting soft-biometric information by protecting biometric data entirely, \ie biometric template protection scheme, as elaborated below:

\begin{LaTeXdescription}
	\item[Cancelable biometrics] \cite{Patel-CancelableBiometrics-2015} obscure biometric signals by applying irreversible transformations to them. To achieve unlinkability, application- or subject-specific transformation parameters, \ie keys, are employed. In case an attacker would be in possession of the key that was used to protect the biometric data, the presented attack could be performed offline. Note that key possession usually does not suffice to revert the protected biometric signal. If the attacker does not have the key, the proposed attack would only be applicable online, provided that a sufficiently large set of face images can be presented to the cancelable biometric system. 
	\item[Biometric cryptosystems] \cite{BUludag04a} do not return biometric comparison scores. In contrast, biometric cryptosystems retrieve keys which are validated and usually only released if these are correct, otherwise a failure message is returned. This means, to perform the proposed attack to a biometric cryptosystem, a certain amount of false matches would need to be achieved when presenting the set of biometric probe images to the system. Obviously, this would depend on the size of the image set the attacker is using and the false match rate the system is operated at. For the conducted experiments, Table~\ref{tab:security-analysis} lists the average proportion of false matches for the best obtained score for decision thresholds corresponding to relevant false match rates in verification mode. It can be observed that for the conducted experiments only extremely low false match rates would considerably reduce the probability of false matches. %Note that, in order to maintain a low false non-match rate, a false match rate of 0.1\% is frequently requested as operation point for facial recognition systems, \eg in \cite{EU-Frontex-BestPracticeABC-2015}. 
	However, if a biometric cryptosystem would return erroneous or random keys in case the key validation fail, the attacker may not be able to correctly identify false matches which in turn would prevent from the proposed attack.
	\item[Homomorphic encryption] \cite{Aguilar-Homomorphic-2013} requires a probe to be encrypted with a public key prior to comparing it to the reference in the encrypted domain. Subsequently, the comparison score is decrypted using the private key. Hence, an attacker would require the  private key of the system in order to obtain comparison scores, which would be a prerequisite to launch the proposed attack. Under the assumption that an attacker has full access to private keys, a direct decryption of encrypted references could be performed. Subsequently, soft-biometric attributes could be reliably extracted from  unprotected references. That is, if the  secrecy of the private keys can be guaranteed in homomorphic encryption schemes, the presented attack can not be applied.   
\end{LaTeXdescription}

\begin{table}[!t]
	\scriptsize
	\centering
	\caption{Relative amount of false matches obtained by the attack in relation to verification-based false match rates (FMRs) (in \%).}
	\label{tab:security-analysis}
    	\begin{adjustbox}{max width=\linewidth}
	\begin{tabular}{cccccc}
	\toprule
	
	\textbf{Method}&\textbf{Attacker}&\textbf{Target}&\textbf{FMR}&\textbf{Threshold}&\textbf{FMs in Attack}\\ 
	%&&&&\textbf{of FMs}\\
	\midrule

	\multirow{12}{*}{PFRNet}        &\multirow{3}{*}{LFW}                  &\multirow{3}{*}{Adience}       &0.001&0.84&0.08\\
	
                                    &                                      &                               &0.01&0.76&1.17\\
                                    
                                    &                                       &                              &0.1&0.61&99.22\\ \cmidrule{2-6}

                                    &\multirow{6}{*}{Adience}               &\multirow{3}{*}{CelebA}       &0.001&0.70&60.93  \\
                                    
                                    &                                      &                               &0.01&0.66&97.66  \\
                                    
                                    &                                       &                              &0.1&0.60&100.00\\ \cmidrule{3-6}

                                    &                                       &\multirow{3}{*}{LFW}          &0.001&0.72&19.98  \\
                                    
                                    &                                      &                               &0.01&0.67&79.97  \\
                                    
                                    &                                      &                               &0.1&0.61&99.90  \\ \cmidrule{2-6}

                                    &\multirow{3}{*}{CelebA}               &\multirow{3}{*}{Adience}       &0.001&0.84&0.01 \\
                                    
                                    &                                      &                               &0.01&0.76&0.73 \\
                                    
                                    &                                      &                               &0.1&0.61&99.51  \\ \cmidrule{1-6}

  \multirow{18}{*}{PE-MIU}          &\multirow{6}{*}{LFW}                   &\multirow{3}{*}{Adience}   &0.001&0.83&0.16\\
                                    
                                    &                                      &                               &0.01&0.70&0.97\\
                                    
                                    &                                       &                              &0.1&0.42&100.00\\ \cmidrule{3-6}

                                    &                                       &\multirow{3}{*}{ColorFeret}   &0.001&0.62&3.42\\
                                    &                                      &                               &0.01&0.47&68.23\\
                                    
                                    &                                       &                              &0.1&0.37&100.00\\ \cmidrule{2-6}

                                    &\multirow{6}{*}{Adience}                   &\multirow{3}{*}{LFW}   &0.001&0.53&56.51\\

                                    &                                      &                               &0.01&0.44&99.95\\
                                    
                                    &                                       &                              &0.1&0.37&100.00\\ \cmidrule{3-6}

                                    &                                      &\multirow{3}{*}{ColorFeret}   &0.001&0.62&11.20\\
                                    
                                    &                                      &                               &0.01&0.47&78.49\\
                                    
                                    &                                       &                              &0.1&0.37&100.00\\ \cmidrule{2-6}

                                    &\multirow{6}{*}{ColorFeret}            &\multirow{3}{*}{LFW}   &0.001&0.53&56.66\\
                                    
                                     &                                      &                               &0.01&0.44&100.00\\
                                    
                                    &                                       &                              &0.1&0.37&100.00\\ \cmidrule{3-6}

                                    &                                        &\multirow{3}{*}{Adience}    &0.001&0.83&0.54\\
                                    
                                    &                                      &                               &0.01&0.70&3.80\\
                                    
                                    &                                       &                              &0.1&0.42&100.00\\

	\bottomrule	
	\end{tabular}
	\end{adjustbox}
\end{table}

In summary, it can be argued that certain template protection mechanisms, in particular biometric cryptosystems and homomorphic encryption, prevent the presented attack while under specific circumstances cancelable biometric systems are expected to be vulnerable to the attack. However, the latter assumption would require further investigations which are beyond the scope of this work.

\subsection{Alternative Attacks}\label{sec:alternative}

Apart from the proposed attack, facial soft–biometric privacy enhancement techniques may be vulnerable to further attacks. As previously mentioned, analyses on privacy protection capabilities of these methods have mostly been conducted by employing well-known machine learning-based classifiers, \eg SVM. However, alternative classification methods based on different (machine learning-based) classifier might be capable of inferring soft-biometric information from privacy-enhanced templates. In addition, classifiers could be trained to retrieve unprotected soft-biometric attributes which are interrelated with a protected soft-biometric attribute. For instance, soft–biometric privacy enhancement methods may be circumvented by deriving gender from another unprotected attribute such as hairstyle or makeup.  

\subsection{Application to other Characteristics}\label{sec:other}

It is worth mentioning that the attack may only be applicable to biometric systems based on characteristics for which the effect of broad homogeneity is observable. It has been shown that biometric attributes can be derived from various popular biometric characteristics \cite{7273870}, \eg fingerprints, iris, or voice. However, this does not necessarily mean that a biometric system based on such characteristics utilises these soft-biometric attributes for recognition purposes. For instance, it has recently been shown that the effect of broad homogeneity can not be observed for commercial iris recognition systems \cite{howard2020quantifying}, while many researchers reported high accuracies for predicting soft-biometric attributes such as gender from iris images \cite{7273870}.

\subsection{Attack Models}\label{sec:models}

Different models exist for describing scenarios and assumptions of attacks on biometric information protection schemes which are standardised in \cite{ISO18-TemplateProtection}. The most restrictive model is referred to as \emph{na{\"i}ve model} in which an adversary has neither information of the underlying algorithm, nor owns a large biometric database. However, it has recently been argued that privacy-enhancing face recognition system should be analysed under Kerckhoffs‘s principle \cite{TerhoerstBIOSIG}. In this \emph{general model}, an adversary is assumed to possess full knowledge of the underlying algorithm. In addition, the adversary may have access to one or more privacy-enhanced templates from one or more databases. The adversary may also possess knowledge of the statistical properties of biometric features. In contrast, in the proposed attack, full knowledge of the underlying algorithm is not required, \ie merely applying it as a black-box is sufficient. More precisely, the attack only requires the privacy-enhancing method as black-box and a small database. It is noteworthy that such a scenario is identical to a scenario in which a machine learning-based classifier would be trained to extract soft-biometric attributes from privacy-enhanced templates. The latter scenario is usually considered in the scientific literature for analysing privacy protection capabilities of soft-biometric privacy enhancement methods \cite{Meden21a}.

\section{Conclusion}
\label{sec:conclusion}
%Some recently proposed privacy-enhancing face recognition methods have been reported to maintain auspicious biometric performance rates while at the same time suppressing or removing soft-biometric attributes at feature level. However, the majority of these works lacks a thorough analysis of privacy protection capabilities with respect to potential attack vectors, a shortcoming that has already been pointed out by earlier work \cite{TerhoerstBIOSIG}. In this context, it is worth noting that efforts have already been made towards a standardised performance testing of biometric information protection \cite{ISO18-TemplateProtection}.

We showed that in order to maintain biometric performance, privacy-enhancing face recognition methods have to retain certain properties of the original face recognition systems. This includes the well-documented effect of broad homogeneity \cite{Howard-DemographicEffectsFace-2019}, \ie face recognition systems produced higher similarity scores for subjects which share certain soft-biometric attributes such as gender or race. Based on these observations an attack was proposed which can be performed offline with the minimal requirements that the algorithm is available as black box along with a small set of arbitrary face images. In experiments, high success rates were achieved for attacking two state-of-the-art algorithms for facial soft-biometric privacy enhancement. Such an attack may also be applicable to other schemes which are conceptually similar to the ones used in the experiments of this work. 
While the proposed attack is applied to infer gender information in this work, it can theoretically be applied to further protected attribute, \eg age or race. It is concluded that the privacy protection capabilities of some facial soft–biometric privacy enhancement techniques are currently over-estimated in published works. Future research on this topic, therefore, needs to focus on more rigorous evaluations when assessing privacy  protection capabilities of soft-biometric privacy-enhancing techniques and consider potential attacks, such as the one introduced in this work.

\section*{Acknowledgements}
\label{sec:acknowledgements}
This work has in part received funding from the European Union’s Horizon 2020  research and innovation programme under the Marie Skłodowska-Curie grant agreement No. 860813 - TReSPAsS-ETN and the German Federal Ministry of Education and Research and the Hessen State Ministry for Higher Education, Research and the Arts within their joint support of the National Research Center for Applied Cybersecurity ATHENE. This work is partially supported by ERCIM, who kindly enabled the internship of P. Terh\"orst at NTNU, Gj{\o}vik, Norway and the ARRS Project J2–1734 “Face deidentification with generative deep models” (FaceGEN).

\bibliographystyle{IEEEtran}
\bibliography{IEEEabrv}

\end{document}